INFSYS
RESEARCH
REPORT

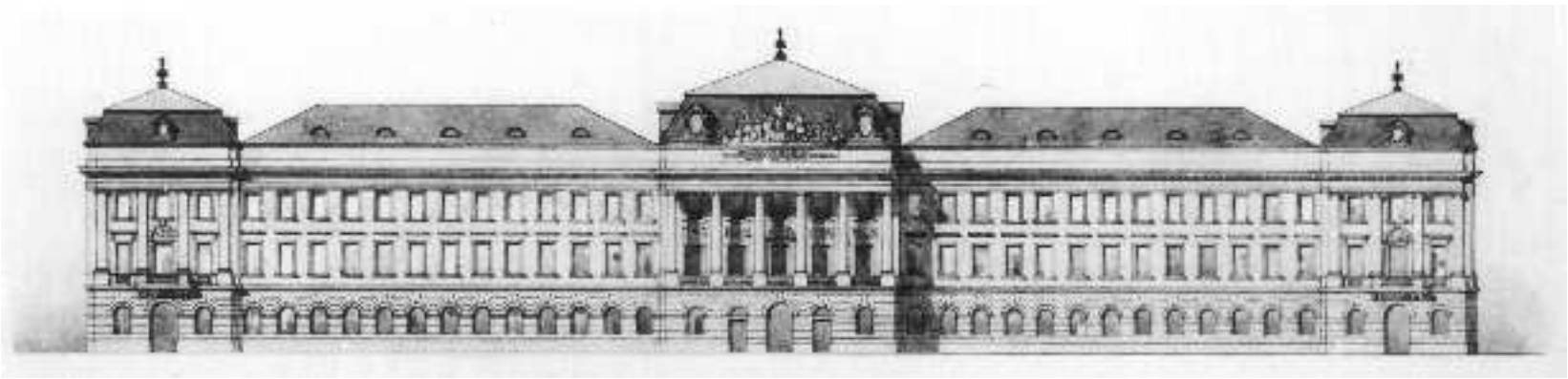

INSTITUT FÜR INFORMATIONSSYSTEME
ABTEILUNG WISSENSBASIERTE SYSTEME

# A LOGIC PROGRAMMING APPROACH TO KNOWLEDGE-STATE PLANNING: SEMANTICS AND COMPLEXITY

Thomas Eiter Wolfgang Faber Nicola Leone

Gerald Pfeifer Axel Polleres

Institut für Informationssysteme
Abtg. Wissensbasierte Systeme
Technische Universität Wien
Favoritenstraße 9-11
A-1040 Wien, Austria
Tel: +43-1-58801-18405
Fax: +43-1-58801-18493
sek@kr.tuwien.ac.at
www.kr.tuwien.ac.at

TECHNISCHE UNIVERSITÄT WIEN

# A LOGIC PROGRAMMING APPROACH TO KNOWLEDGE-STATE PLANNING: SEMANTICS AND COMPLEXITY

Thomas Eiter[1], Wolfgang Faber[1], Nicola Leone[2], Gerald Pfeifer[3], Axel Polleres[1]

**Abstract.** We propose a new declarative planning language, called $\mathcal{K}$, which is based on principles and methods of logic programming. In this language, transitions between states of knowledge can be described, rather than transitions between completely described states of the world, which makes the language well-suited for planning under incomplete knowledge. Furthermore, it enables the use of default principles in the planning process by supporting negation as failure. Nonetheless, $\mathcal{K}$ also supports the representation of transitions between states of the world (i.e., states of complete knowledge) as a special case, which shows that the language is very flexible. As we demonstrate on particular examples, the use of knowledge states may allow for a natural and compact problem representation. We then provide a thorough analysis of the computational complexity of $\mathcal{K}$, and consider different planning problems, including standard planning and secure planning (also known as *conformant planning*) problems. We show that these problems have different complexities under various restrictions, ranging from NP to NEXPTIME in the propositional case. Our results form the theoretical basis for the $\texttt{DLV}^{\mathcal{K}}$ system, which implements the language $\mathcal{K}$ on top of the DLV logic programming system.

**Keywords:** Answer sets, conformant planning, computational complexity, declarative planning, incomplete information, knowledge-states, secure planning

---

[1]Institut für Informationssysteme, Abteilung Wissensbasierte Systeme, Technische Universität Wien, Favoritenstraße 9-11, A-1040 Vienna, Austria. E-mail: {eiter, faber, axel}@kr.tuwien.ac.at.

[2]Department of Mathematics, University of Calabria, 87030 Rende (CS), Italy. E-mail: leone@unical.it.

[3]Institut für Informationssysteme, Abteilung Datenbanken und AI, Technische Universität Wien, Favoritenstraße 9-11, A-1040 Vienna, Austria. E-mail: {pfeifer}@dbai.tuwien.ac.at.

**Acknowledgements**: This work was supported by FWF (Austrian Science Funds) under the projects P14781-INF, and Z29-INF.

Preliminary results of this paper appeared in “Planning under Incomplete Knowledge,” *Proceedings of the First International Conference on Computational Logic (CL 2000), London, UK, July 24–28*, J.W. Lloyd et al., editors, Lecture Notes in Computer Science 1861, Springer, 2000, pp. 807–821.

# Contents

# 1 Introduction

Since intelligent agents must have planning capabilities, planning has been an important problem in AI since its very beginning, and numerous approaches and methods have been developed in extensive work over the last decades. The formulation of planning as a problem in logic dates back to a proposal of McCarthy in the 1950s; the breakthrough of Robinson's resolution method laid the basis for deductive planning as in Green's paper [31] and the well-known situation calculus [51]. However, because of defects such as the well-known frame problem, deductive planning lost attention, while the STRIPS approach [20], a hybrid between logic and procedural computation, and its derivates were gaining importance. For a long period then, fairly no other logic-related planning systems were explored.

In the last 12 years, however, logic-based planning celebrated a renaissance, emerging in different streams of work:

- Solutions to the frame problem have been worked out, and deductive planning based on the situational calculus has been considered extensively, in particular by the Toronto group, leading to the GOLOG planning language [40]. In parallel, planning in the event calculus [38] has been pursued, starting from [15, 63].

- Formulating planning problems as logical satisfiability problems, proposed by Kautz and Selman [36], enabled to solve large planning problems which could not be solved by specialized planning systems, and led to the efficient Blackbox planning system [37]. In the same spirit, other approaches reduced planning problems to computational tasks in logical formalisms, including logic programming [8, 65], model checking [5, 6], and Quantified Boolean Formulas [60].

- Planning as a task in logic-based languages for reasoning about actions, which were developed in the context of logics for knowledge representation and logic programming, e.g. [23, 35, 26, 27, 28, 34, 48, 67]; see [24, 68] for surveys. Implementing these languages using, in the spirit of Kautz and Selman, satisfiability solvers led to the causal calculator (CCALC) [49, 47] and the $\mathcal{C}$-plan system [25], which is based on the important $\mathcal{C}$ action language [27].

In very influential papers, Lifschitz proposed answer set programming as a tool for problem solving, and in particular for planning [43, 44]. In this approach, planning problems, formulated in a domain-independent planning language, are mapped into an extended logic program such that the answer sets of this program give the solutions of the planning problem (cf. also [45]). In this way, planners may be created which support expressive action description languages and, by the use of efficient answer sets engines such as smodels [33] or `DLV` [13], allow for efficient problem solving.

In our work, we pursue this suggestion and develop it further. In the present paper, we propose a new language, $\mathcal{K}$, for planning under incomplete knowledge. We name it $\mathcal{K}$ to emphasize that it describes transitions between *states of knowledge* rather than between *states of the world*. Namely, language $\mathcal{C}$ and many others are based on extensions of classical logics and describe transitions between *possible states of the world*. Here, a state of the world is characterized by the truth values of a number of fluents, i.e., predicates describing relevant properties of the domain of discourse, where every fluent necessarily is either true or false. An action is applicable only if some precondition (formula over the fluents) is true in the current state, and executing this action changes the current state by modifying the truth values of some fluents.

However, planning agents usually don't have a *complete* view of the world. Even if their knowledge is incomplete, that is, a number of fluents is unknown, they must take decisions, execute actions, and reason on the basis of their (incomplete) information at hand. For example, imagine a robot in front of a door. If it is unknown whether the door is open, the robot may decide to push back. Alternatively, it might decide to sense the door status in order to obtain complete information. However, this requires that a suitable sensing action is available and, importantly, actually executable (that is, the sensor is not broken). Thus, even in the presence of sensing, some fluents may remain unknown and leave an agent in a state of incomplete information.

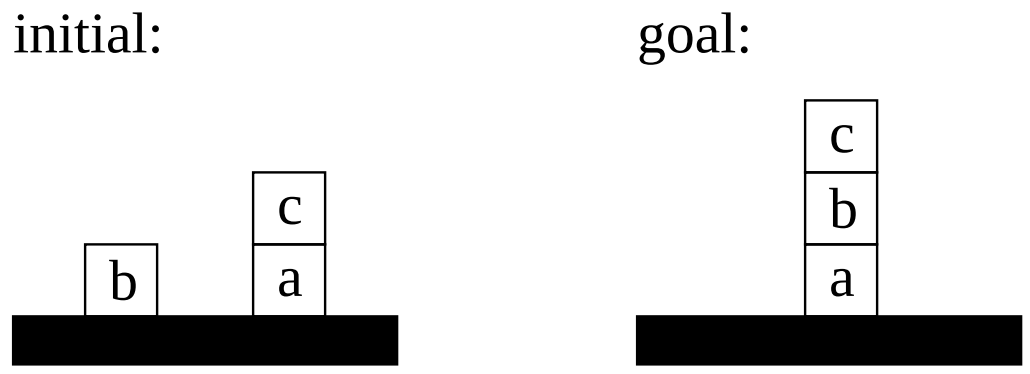

Figure 1: A blocksworld example.

Our language $\mathcal{K}$ adopts a three-valued view of fluents in which their values might be true, false, or unknown. The language is very flexible, and is capable of modeling transitions between states of the world (i.e., states of complete knowledge) and of reasoning about them as a particular case, as we shall discuss. Compared to similar planning languages, $\mathcal{K}$ is closer in spirit to answer set semantics [22] than to classical logics. It allows for the use of default negation, exploiting the power of answer sets to deal with incomplete knowledge. We also analyze the computational complexity of $\mathcal{K}$, which provides the theoretical background for the DLV$^{\mathcal{K}}$ system implementing $\mathcal{K}$ on top of the DLV system [13, 16]. DLV$^{\mathcal{K}}$ provides a powerful declarative planning system, which is ready-to-use for experiments (see `<URL:http://www.dbai.tuwien.ac.at/proj/dlv/>`).

### 1.1 A Brief Overview of $\mathcal{K}$

As an appetizer, we give a brief exposition of the main features of the language $\mathcal{K}$, which will be formally defined in Section 2. We occasionally refer to well-known planning problems in the "blocksworld" domain, which require turning given configurations of blocks into goal configurations (see Figure 1).

**Background Knowledge** The planning domain has a background which is represented by a normal (that is, disjunction-free) stratified logic program. The rules and facts of this program define predicates which are not subject to change, i.e. represent *static* knowledge. An example in blocksworld is `block(B)`, which states the (unchangeable) property that `B` is a block.

**Type Declarations** The ranges of the arguments of fluents and actions are typed, by stating that certain predicates must hold on them. For example,

$$\texttt{move(B,L) requires block(B), location(L).}$$

specifies the types for the arguments of action `move`. The literals after the `requires` keyword (here, `block(B)` and `location(L)`) must be positive literals of the static background knowledge mentioned above.

**Causation Rules** The main construct of $\mathcal{K}$ are *causation rules*. They are syntactically similar to rules of the language $\mathcal{C}$ [27, 43, 45] and have the form:

caused f if B after A.

Intuitively, this rule reads "If B is known to be true in the current state and A is known to be true in the previous state, then f is known to be true in the current state." Both the if part and the after part may be empty (which means that it is true).

**Negation** Default (or weak) negation "not" can be used in the if and the after part of the rules. It allows for natural modeling of inertial properties, default properties, and dealing with incomplete knowledge in general, similar to logic programming with answer set semantics. Furthermore, strong negation ("$\neg$", written in programs as "$-$") is supported as well. In order to support convenient problem representation, $\mathcal{K}$ provides several constructs, which are "implemented" through weak negation, as, e.g., inertial on(X, Y).

which informally states that on(X, Y) is concluded to hold in the current state if on(X, Y) held at the previous state and $-$on(X, Y) is not explicitly known to hold, or default $-$ on(X, Y).

which states that $-$on(X, Y) is concluded to hold unless on(X, Y) is known to hold (as it has been explicitly entailed by some causation rule).

**Executability of Actions** In order to be eligible for execution, any action needs to satisfy some precondition in a given state of knowledge, which can be stated using executability statements. For example, executable move(X, Y) if not occupied(X), not occupied(Y), X <> Y.

states that block X can be moved on location Y if both X and Y are clear and X $\neq$ Y (assuming proper typing). Multiple executability statements for the same action are allowed. If the body is empty, it means that the action always qualifies for execution, provided that the type restrictions on X and Y are respected. On the other hand, execution of an action A under condition B can also be blocked, by the statement nonexecutable A if B.

In case of conflicts, nonexecutable A overrides executable A.

**Integrity Constraints** In general, a causation rule expresses a state constraint that must be fulfilled in all states. It is very common to state *integrity constraints* for states (possibly referring to the respective preceding state), i.e., conjunctions of literals which can not simultaneously be satisfied. To facilitate convenient representation of integrity constraints, $\mathcal{K}$ provides a statement

forbidden B after A

as a shortcut for caused false if B after A. Intuitively, it discards any state where B is (known to be) true, if A is (known to be) true in the previous state.

**Initial State Constraints** $\mathcal{K}$ allows to declare causation rules with empty `after`-part that should apply to the initial state only. Such rules, which represent constraints on the initial state, must be preceded by the keyword "`initially` :". For example,

$$\texttt{initially}: \ \texttt{forbidden}\ \texttt{block}(\texttt{B}),\ \texttt{not}\ \texttt{supported}(\texttt{B}).$$

requires that the fluent `supported` is true for every block in the initial state; the constraint is irrelevant for all subsequent states. Initial state constraints may profitably reduce computation effort: If we are guaranteed that actions preserve some property $P$, then it is sufficient to check the validity of $P$ only on the initial state to ensure that it holds in any state.

**Parallel Execution of Actions** By default, simultaneous execution of actions is allowed in $\mathcal{K}$. This can be prohibited by suitable rules; however, for the user's convenience, a statement

$$\texttt{noConcurrency}.$$

is provided as a shortcut which enforces the execution of at most one action at a time.

**Handling of Complete and Incomplete Knowledge** $\mathcal{K}$ also allows one to represent transitions between possible states of the world (which can be seen as states of complete knowledge). First of all, we can easily enforce that the knowledge on some fluent `f` is complete, using a rule

$$\texttt{forbidden}\ \texttt{not}\ \texttt{f},\ \texttt{not}\ -\texttt{f}.$$

Moreover, we can "totalize" the knowledge of a fluent by declaring

$$\texttt{total}\ \texttt{f}.$$

which means that, unless a truth value for `f` can be derived, the cases where `f` resp. $-$`f` is true will be both considered. In other words, every state will be "totalized" by adding `f` or $-$`f`, if none of them is true.

**Goals and Plans** A goal is a conjunction of ground literals; a plan for a goal is a sequence of (in general, sets of) actions whose execution leads from an initial state to a state where all literals in the goal are true. In $\mathcal{K}$, the goal is followed by a question mark and by the number of allowed steps in a plan. For instance,

$$\texttt{on}(\texttt{c},\texttt{b}),\ \texttt{on}(\texttt{b},\texttt{a})\ ?\ (3)$$

requests a plan of length 3 for the goal of Figure 1.

This concludes the exposition of the $\mathcal{K}$ planning language. We remark at this point that the $\text{DLV}^{\mathcal{K}}$ planning system contains the command

$$\texttt{securePlan}.$$

by which we can ask the system to compute only *secure plans* (often called *conformant plans* or *fail-safe plans* in the literature [29, 64]). Informally, a plan is secure, if it is applicable starting at any legal initial state, and enforces the goal, regardless of how the state evolves. Using this feature, we can also model *possible-worlds planning with an incomplete initial state*, where the initial world is only partially known, and we are looking for a plan reaching the desired goal from every possible world according to the initial state. Note that, by our complexity results, unlike the other statements above the "`securePlan`." command can *not* be expressed as a shortcut in language $\mathcal{K}$, and thus has to be realized at an external level.

### 1.2 Contributions

The main contributions of the present paper are the following:

(**1**) We propose a new planning language, called $\mathcal{K}$, which is based on logic programming. We formally define language $\mathcal{K}$ and provide a declarative, model theoretic semantics for it. Importantly, the language supports also default (nonmonotonic) negation, which enriches the knowledge modeling power of $\mathcal{K}$. To capture the intuitive meaning of default negation, the formal semantics of the planning language $\mathcal{K}$ is given in two steps like for stable models in logic programming [22].

(**2**) We illustrate the knowledge modeling features of the language by encoding some classical planning problems in $\mathcal{K}$, in particular different versions of blocksworld and "bomb in the toilet" planning problems [52]. We proceed incrementally, presenting all main features of $\mathcal{K}$ and their usage for knowledge representation and reasoning in planning domains. In the course of this, we show $\mathcal{K}$ encodings of classical planning problems (dealing with complete knowledge), and we further describe how conformant planning problems (in presence of incomplete knowledge on the initial state, or in presence of nondeterministic action effects) can be encoded in $\mathcal{K}$.

As we show, the language $\mathcal{K}$ is capable of expressing classical encodings based on states of the world. However, by its design it is very well-suited for encodings based on states of knowledge. We show both types of encodings on some "bomb in the toilet" planning problems, and discuss the two different approaches, highlighting some computational advantages of the encodings based on states of knowledge.

(**3**) We perform a thorough study of the complexity of major planning problems in the language $\mathcal{K}$, where we focus on the propositional case. (Results for the first-order case can be obtained in the usual manner.) In particular, we consider the problems of deciding the existence of an optimistic (i.e., standard) plan for a given length, the problem of checking whether such a plan is secure (i.e., conformant), and the combined problem of finding a secure (i.e., conformant) plan, under various restrictions on the planning instances. For formal definitions of optimistic and secure plans, we refer to Section 2.2.

It appears that deciding the existence of an optimistic plan achieving the goal in a fixed number of steps is NP-complete, while it is PSPACE-complete in general. Thus, in general we have the same complexity as for planning in corresponding STRIPS-like systems [20], which are well-known PSPACE-complete [3]. On the other hand, finding secure plans is obviously harder, because it allows us to encode also planning under incomplete initial states as in [1], which was shown to be $\Sigma_2^P$-complete there for polynomial-length plans. In fact, deciding the existence of a secure plan of variable (arbitrary) length is NEXPTIME-complete, and thus not polynomially reducible to planning in STRIPS-like systems or to QBF-solvers, which can only express problems in PSPACE (unless NEXPTIME collapses to PSPACE). Even under fixed plan length, this problem is $\Sigma_3^P$-complete, and thus rather complex; further restrictions have to be imposed to lower its complexity. To this end, we introduce meaningful subclasses of planning domains and problems, in particular *proper* and *plain* planning domains resp. problems. As we show, for proper planning domains, existence of a secure plan having a fixed number of steps is only mildly harder than NP if concurrent actions are not allowed.

Our complexity results give a clear picture of the feasibility of polynomial-time translations for particular planning problems into computational logic systems such as Blackbox [37], CCALC [47], smodels [33], DLV, satisfiability checkers, e.g. [2, 74], or Quantified Boolean Formula (QBF) solvers [4, 61, 18].

### 1.3 Structure of the Paper

The rest of the paper is structured as follows. The next section formally introduces the language $\mathcal{K}$, and provides the syntax and formal semantics of the core language, as well as enhancements of the language by macro constructs that are useful "syntactic sugar" for conveniently representing problems. After that, we consider in Section 3 knowledge representation in $\mathcal{K}$, where different aspects such as planning with incomplete initial states, representation of nondeterministic action effects, and knowledge-based encodings of the latter are discussed. In Section 4 we then embark on our study of the complexity of language $\mathcal{K}$, and present an overview of the problems we considered and the main results that we obtained. Section 5 is then devoted to the derivation of these complexity results. In Section 6, we discuss related work, and the final Section 7 discusses further work and draws some conclusions.

The present paper is part I in a series of papers which comprehensively describe our work, and contains the foundational semantic definitions and theoretical results; part II [12] reports about the $\texttt{DLV}^{\mathcal{K}}$ system (which is freely available at `<URL:http://www.dbai.tuwien.ac.at/proj/dlv/>`) and in particular contains an experimental evaluation and comparisons to other planning systems (for a theoretical account, see also Section 6).

## 2 Language $\mathcal{K}$

In this section, we will detail syntax and semantics of the language $\mathcal{K}$ that we have briefly introduced in the previous section.

### 2.1 Basic Syntax

#### 2.1.1 Actions, Fluents, and Types

Let $\sigma^{act}$, $\sigma^{fl}$, and $\sigma^{typ}$ be disjoint sets of action, fluent and type names, respectively. These names are effectively predicate symbols with associated arity ($\geq 0$). Here, $\sigma^{fl}$ and $\sigma^{act}$ are used to describe *dynamic knowledge*, whereas $\sigma^{typ}$ is used to describe *static background knowledge*. Furthermore, let $\sigma^{con}$ and $\sigma^{var}$ be the disjoint sets of constant and variable symbols, respectively.

**Definition 2.1** For $p \in \sigma^{act}$ (resp. $\sigma^{fl}$, $\sigma^{typ}$), an *action (resp. fluent, type) atom* is defined as $p(t_1, \ldots, t_n)$, where $n$ is the arity of $p$ and $t_1, \ldots, t_n \in \sigma^{con} \cup \sigma^{var}$. An action (resp. fluent, type) literal is an action (resp. fluent, type) atom $a$ or its negation $\neg a$, where "$\neg$" is the true negation symbol, for which we also use the customary "–".

As usual, a literal (and any other syntactic object) is *ground*, if it does not contain variables.

Given a literal $l$, let $\neg.l$ denote its complement, i.e., $\neg.l = a$ if $l = \neg a$ and $\neg.l = \neg a$ if $l = a$, where $a$ is an atom. A set $L$ of literals is *consistent*, if $L \cap \neg.L = \emptyset$. Furthermore, $L^+$ (resp., $L^-$) denotes the set of positive (resp., negative) literals in $L$.

The set of all action (resp. fluent, type) literals is denoted as $\mathcal{L}_{act}$ (resp. $\mathcal{L}_{fl}$, $\mathcal{L}_{typ}$). Furthermore, $\mathcal{L}_{fl,typ} = \mathcal{L}_{fl} \cup \mathcal{L}_{typ}$; $\mathcal{L}_{dyn} = \mathcal{L}_{fl} \cup \mathcal{L}^+_{act}$ ($dyn$ stands for *dynamic literals*); and $\mathcal{L} = \mathcal{L}_{fl,typ} \cup \mathcal{L}^+_{act}$. [1]

All actions and fluents must be declared using statements as follows.

[1] Note that this definition only allows positive action literals.

**Definition 2.2** An *action* (resp., *fluent*) *declaration*, is of the form:

$$p(X_1, \ldots, X_n)\ \texttt{requires}\ t_1, \ldots, t_m \tag{1}$$

where $p \in \mathcal{L}^+_{act}$ (resp. $p \in \mathcal{L}^+_{fl}$), $X_1, \ldots, X_n \in \sigma^{var}$ where $n \geq 0$ is the arity of $p$, $t_1, \ldots, t_m \in \mathcal{L}_{typ}$, $m \geq 0$, and every $X_i$ occurs in $t_1, \ldots, t_m$.

If $m = 0$, the keyword `requires` may be omitted.

We next define causation rules, by which static and dynamic dependencies of fluents on other fluents and actions are specified.

**Definition 2.3** A *causation rule* (*rule*, for short) is an expression of the form

$$\begin{array}{l} \texttt{caused}\ f\ \texttt{if}\ b_1, \ldots, b_k, \texttt{not}\ b_{k+1}, \ldots, \texttt{not}\ b_l \\ \qquad\qquad \texttt{after}\ a_1, \ldots, a_m, \texttt{not}\ a_{m+1}, \ldots, \texttt{not}\ a_n \end{array} \tag{2}$$

where $f \in \mathcal{L}_{fl} \cup \{\texttt{false}\}$, $b_1, \ldots, b_l \in \mathcal{L}_{fl,typ}$, $a_1, \ldots, a_n \in \mathcal{L}$, $l \geq k \geq 0$, and $n \geq m \geq 0$.

Rules where $n = 0$ are referred to as *static rules*, all other rules as *dynamic rules*. When $l = 0$, the keyword `if` is omitted; likewise, if $n = 0$, the keyword `after` is dropped. If both $l = n = 0$ then `caused` is optional.

To access the parts of a causation rule $r$, we use the following notations: $\mathsf{h}(r) = \{f\}$, $\mathsf{post}^+(r) = \{b_1, \ldots, b_k\}$, $\mathsf{post}^-(r) = \{b_{k+1}, \ldots, b_l\}$, $\mathsf{pre}^+(r) = \{a_1, \ldots, a_m\}$, $\mathsf{pre}^-(r) = \{a_{m+1}, \ldots, a_n\}$, and $\mathsf{lit}(r) = \{f, b_1, \ldots, b_l, a_1, \ldots, a_n\}$. Intuitively, $\mathsf{pre}^+(r)$ accesses the state before some action(s) happen, and $\mathsf{post}^+(r)$ the part after the actions have been executed.

While the scope of general static rules is over all knowledge states, it is often useful to specify rules only for the initial states.

**Definition 2.4** An *initial state constraint* is a static rule of form (2) preceded by the keyword `initially`.

The language $\mathcal{K}$ allows STRIPS-style [20] conditional execution of actions, where $\mathcal{K}$ allows several alternative executability conditions for an action which is beyond the repertoire of standard STRIPS.

**Definition 2.5** An *executability condition* is an expression of the form

$$\texttt{executable}\ a\ \texttt{if}\ b_1, \ldots, b_k, \texttt{not}\ b_{k+1}, \ldots, \texttt{not}\ b_l \tag{3}$$

where $a \in \mathcal{L}^+_{act}$ and $b_1, \ldots, b_l \in \mathcal{L}$, and $l \geq k \geq 0$.

If $l = 0$ (which means that the executability is unconditional), then the keyword `if` is skipped.

Given an executability condition $e$, we access its parts with $\mathsf{h}(e) = \{a\}$, $\mathsf{pre}^+(e) = \{b_1, \ldots, b_k\}$, $\mathsf{pre}^-(e) = \{b_{k+1}, \ldots, b_l\}$, and $\mathsf{lit}(e) = \{a, b_1, \ldots, b_l\}$. Intuitively, $\mathsf{pre}^-(e)$ refers to the state at which some action's suitability is evaluated. Here, as opposed to causation rules we do not consider a state after the execution of actions, and so no part $\mathsf{post}^+(r)$ is needed. Nonetheless, for convenience we define $\mathsf{post}^+(e) = \mathsf{post}^-(e) = \emptyset$.

Furthermore, for any executability condition, a rule, or an initial state constraint $r$, we define $\mathsf{post}(r) = \mathsf{post}^+(r) \cup \mathsf{post}^-(r)$, $\mathsf{pre}(r) = \mathsf{pre}^+(r) \cup \mathsf{pre}^-(r)$, and $\mathsf{b}(r) = \mathsf{b}^+(r) \cup \mathsf{b}^-(r)$, where $\mathsf{b}^+(r) = \mathsf{post}^+(r) \cup \mathsf{pre}^+(r)$, and $\mathsf{b}^-(r) = \mathsf{post}^-(r) \cup \mathsf{pre}^-(r)$.

**Example 2.1** Consider the following type declarations, causation rule, and executability condition, respectively, where $\sigma^{typ} = \{\mathtt{r}, \mathtt{s}\}$, $\sigma^{fl} = \{\mathtt{f}\}$, and $\sigma^{act} = \{\mathtt{ac}\}$:

$$\begin{array}{ll} t_1: & \mathtt{f(X)\ requires\ -r(X,Y),\ s(Y,Y).} \\ t_2: & \mathtt{ac(X,Y)\ requires\ s(X,Y).} \\ r_1: & \mathtt{f(X)\ if\ s(X,X),\ not\ -f(X)\ after\ ac(X,Y),\ not\ -r(X,X).} \\ e_1: & \mathtt{executable\ ac(X,Y)\ if\ s(Z,Y),\ not\ f(X),\ Z <> Y.} \end{array}$$

Then, we have $\mathsf{h}(r_1) = \{\mathtt{f(X)}\}$, $\mathsf{pre}(r_1) = \{\mathtt{ac(X,Y)}, -\mathtt{r(X,X)}\}$ and $\mathsf{post}(r_1) = \{\mathtt{s(X,X)}, -\mathtt{f(X)}\}$. Furthermore, $\mathsf{h}(e_1) = \mathtt{ac(X,Y)}$ and $\mathsf{pre}(e_1) = \{\mathtt{s(Z,Y)}, \mathtt{f(X)}, \mathtt{Z} <> \mathtt{Y}\}$; here the inequality predicate $\mathtt{Z} <> \mathtt{Y}$ is regarded as default negation $\mathtt{not}\ (\mathtt{Z} = \mathtt{Y})$, where equality "=" is a built-in which is tacitly present in $\sigma^{typ}$.

### 2.1.2 Safety Restriction

All rules (including initial state constraints and executability conditions) have to satisfy the following syntactic restriction, which is similar to the notion of safety in logic programs [70]. All variables in a default-negated type literal must also occur in some literal which is not a default-negated type literal.

Thus, safety is required only for variables appearing in default-negated type literals, while it is not required at all for variables appearing in fluent and action literals. The reason is that the range of the latter variables is implicitly restricted by the respective type declarations. Observe that the rules in Example 2.1 are all safe.

### 2.1.3 Planning Domains and Planning Problems

We now define planning domains and problems. Let us call any pair $\langle D, R\rangle$ where $D$ is a finite set of action and fluent declarations and $R$ is a finite set of safe causation rules, safe initial state constraints, and safe executability conditions, an *action description*.

**Definition 2.6** A *planning domain* is a pair $PD = \langle \Pi, AD\rangle$, where $\Pi$ is a normal stratified Datalog program (referred to as *background knowledge*), which is assumed to be safe in the standard LP sense (cf. [70]), and $AD$ is an action description. We say that $PD$ is *positive*, if no default negation occurs in $AD$.

Planning domains represent the universe of discourse for solving concrete planning problems, which are defined next.

**Definition 2.7** A *planning problem* $\mathcal{P} = \langle PD, q\rangle$ is a pair of a planning domain $PD$ and a query $q$, where a *query* is an expression of the form

$$g_1, \ldots, g_m, \mathtt{not}\ g_{m+1}, \ldots, \mathtt{not}\ g_n\ ?\ (i) \tag{4}$$

where $g_1, \ldots, g_n \in \mathcal{L}_{fl}$ are variable-free, $n \geq m \geq 0$, and $i \geq 0$ denotes the plan length.

## 2.2 Semantics

For defining the semantics of $\mathcal{K}$ planning domains and planning problems, we start with the preliminary definition of the typed instantiation of a planning domain. This is similar to the grounding of a logic program, with the difference being that only correctly typed fluent and action literals are generated.

### 2.2.1 Typed Instantiation

Let substitutions and their application to syntactic objects be defined as usual (i.e., assignments of constants to variables which replace the variables throughout the objects).

Let $PD = \langle \Pi, \langle D, R \rangle \rangle$ be a planning domain, and let $M$ be the (unique) answer set of $\Pi$ [22]. Then, $\theta(p(X_1, \ldots, X_n))$ is a *legal action* (resp. *fluent*) *instance* of an action (resp. fluent) declaration $d \in D$ of the form (1), if $\theta$ is a substitution defined over $X_1, \ldots, X_n$ such that $\{\theta(t_1), \ldots, \theta(t_m)\} \subseteq M$. By $\mathcal{L}_{PD}$ we denote the set of all legal action and fluent instances.

Based on this, we now define the instantiation of a planning domain respecting type information as follows.

**Definition 2.8** For any planning domain $PD = \langle \Pi, \langle D, R \rangle \rangle$, its *typed instantiation* is given by $PD{\downarrow} = \langle \Pi{\downarrow}, \langle D, R{\downarrow} \rangle \rangle$, where $\Pi{\downarrow}$ is the grounding of $\Pi$ (over $\sigma^{con}$) and $R{\downarrow} = \{\theta(r) \mid r \in R,\ \theta \in \Theta_r\}$, where $\Theta_r$ is the set of all substitutions $\theta$ of the variables in $r$ using $\sigma^{con}$, such that $\mathsf{lit}(\theta(r)) \cap \mathcal{L}_{dyn} \subseteq \mathcal{L}_{PD} \cup (\neg.\mathcal{L}_{PD} \cap \mathcal{L}_{fl}^{-})$.

In other words, in $PD{\downarrow}$ we replace $\Pi$ and $R$ by their ground versions, but keep of the latter only rules where the atoms of all fluent and action literals agree with their declarations. We say that a $PD = \langle \Pi, \langle D, R \rangle \rangle$ is *ground*, if $\Pi$ and $R$ are ground, and moreover that it is *well-typed*, if $PD$ and $PD{\downarrow}$ coincide.

### 2.2.2 States and Transitions

We are now prepared to define the semantics of a planning domain, which is given in terms of states and transition between states.

**Definition 2.9** A *state* with respect to a planning domain $PD$ is any consistent set $s \subseteq \mathcal{L}_{fl} \cap (\mathsf{lit}(PD) \cup \mathsf{lit}(PD)^{-})$ of legal fluent instances and their negations. A tuple $t = \langle s, A, s' \rangle$ where $s, s'$ are states and $A \subseteq \mathcal{L}_{act} \cap \mathsf{lit}(PD)$ is a set of legal action instances in $PD$ is called a *state transition*.

Observe that a state does not necessarily contain either $f$ or $\neg f$ for each legal instance $f$ of a fluent. In fact, a state may even be empty ($s = \emptyset$). The empty state represents a “tabula rasa” state of knowledge about the fluent values in the planning domain. Furthermore, in this definition, state transitions are not constrained – this will be done in the definition of legal state transitions, which we develop now. To ease the intelligibility of the semantics, we proceed in analogy to the definition of answer sets in [22] in two steps. We first define the semantics for positive planning problems, i.e., planning problems without default negation, and then we define the semantics of general planning domains by a reduction to positive planning domains.

In what follows, we assume that $PD = \langle \Pi, \langle D, R \rangle \rangle$ is a ground planning domain which is well-typed, and that $M$ is the unique answer set of $\Pi$. For any other $PD$, the respective concepts are defined through its typed grounding $PD{\downarrow}$.

**Definition 2.10** A state $s_0$ is a *legal initial state* for a positive $PD$, if $s_0$ is the smallest (under inclusion) set such that $\mathsf{post}(c) \subseteq s_0 \cup M$ implies $\mathsf{h}(c) \subseteq s_0$, for all initial state constraints and static rules $c \in R$.

For a positive $PD$ and a state $s$, a set $A \subseteq \mathcal{L}_{act}^{+}$ is called *executable action set* w.r.t. $s$, if for each $a \in A$ there exists an executability condition $e \in R$ such that $\mathsf{h}(e) = \{a\}$, $\mathsf{pre}(e) \cap \mathcal{L}_{fl,typ} \subseteq s \cup M$, and $\mathsf{pre}(e) \cap \mathcal{L}_{act}^{+} \subseteq A$. Note that this definition allows for modeling dependent actions, i.e. actions which depend on the execution of other actions.

**Definition 2.11** Given a positive $PD$, a state transition $t = \langle s, A, s' \rangle$ is called *legal*, if $A$ is an executable action set w.r.t. $s$ and $s'$ is the minimal consistent set that satisfies all causation rules w.r.t. $s \cup A \cup M$. That is, for every causation rule $r \in R$, if (i) $\mathsf{post}(r) \subseteq s' \cup M$, (ii) $\mathsf{pre}(r) \cap \mathcal{L}_{fl,typ} \subseteq s \cup M$, and (iii) $\mathsf{pre}(r) \cap \mathcal{L}_{act} \subseteq A$ all hold, then $\mathsf{h}(r) \neq \{\mathtt{false}\}$ and $\mathsf{h}(r) \subseteq s'$.

The above definitions are now generalized to a well-typed ground $PD$ containing default negation by means of a reduction to a positive planning domain, which is similar in spirit to the Gelfond-Lifschitz reduction [22]:

**Definition 2.12** Let $PD$ be a ground and well-typed planning domain, and let $t = \langle s, A, s' \rangle$ be a state transition. Then, the *reduction* $PD^t = \langle \Pi, \langle D, R^t \rangle \rangle$ of $PD$ by $t$ is the planning domain where $R^t$ is obtained from $R$ by deleting

1. every $r \in R$, for which either $\mathsf{post}^-(r) \cap (s' \cup M) \neq \emptyset$ or $\mathsf{pre}^-(r) \cap (s \cup A \cup M) \neq \emptyset$ holds, and
2. all default literals $\mathtt{not}\ L$ ($L \in \mathcal{L}$) from the remaining $r \in R$.

Note that $PD^t$ is positive and ground. Legal initial states, executable action sets, and legal state transitions are now defined as follows.

**Definition 2.13** Let $PD$ be any planning domain. Then, a state $s_0$ is a *legal initial state*, if $s_0$ is a legal initial state for $PD^t$, where $t = \langle \emptyset, \emptyset, s_0 \rangle$; a set $A$ is an *executable action set* in $PD$ w.r.t. a state $s$, if $A$ is executable w.r.t. $s$ in $PD^t$ with $t = \langle s, A, \emptyset \rangle$; and, a state transition $t = \langle s, A, s' \rangle$ is *legal* in $PD$, if it is legal in $PD^t$.

**Example 2.2** Reconsider the type declarations $t_1$ and $t_2$, causation rule $r_1$ and executability condition $e_1$ in Example 2.1. Suppose $\sigma^{con}$ contains two constants $\mathtt{a}$ and $\mathtt{b}$, and that the background knowledge $\Pi$ has the following answer set: $M = \{-\mathtt{r}(\mathtt{a},\mathtt{b}), \mathtt{r}(\mathtt{b},\mathtt{a}), \mathtt{s}(\mathtt{a},\mathtt{a}), \mathtt{s}(\mathtt{a},\mathtt{b}), \mathtt{s}(\mathtt{b},\mathtt{b})\}$. Then, e.g. $\mathtt{f}(\mathtt{a})$ is a legal fluent instance of $t_1$,

$$\mathtt{f}(\mathtt{X})\ \ \mathtt{requires}\ \ -\mathtt{r}(\mathtt{X},\mathtt{Y}),\ \ \mathtt{s}(\mathtt{Y},\mathtt{Y}).$$

where $\theta = \{\mathtt{X} = \mathtt{a}, \mathtt{Y} = \mathtt{b}\}$. Similarly, $\mathtt{ac}(\mathtt{a},\mathtt{b})$ is a legal action instance of declaration $t_2$,

$$\mathtt{ac}(\mathtt{X},\mathtt{Y})\ \ \mathtt{requires}\ \ \mathtt{s}(\mathtt{X},\mathtt{Y}).$$

where $\theta = \{\mathtt{X} = \mathtt{a}, \mathtt{Y} = \mathtt{b}\}$. Thus, $\mathtt{f}(\mathtt{a})$ and $\mathtt{ac}(\mathtt{a},\mathtt{b})$ belong to $\mathcal{L}_{PD}$. The empty set $s_0 = \emptyset$ is a legal initial state, and in fact the only one since there are no causation rules which apply to initial states in $PD$, and thus also not in $PD^t$ for every $t$. The action set $A = \{\mathtt{ac}(\mathtt{a},\mathtt{b})\}$ is executable w.r.t. $s_0$, since for $t = \langle s_0, A, \emptyset \rangle$, the reduct $PD^t$ contains the executability condition

$$e_1' :\ \ \mathtt{executable}\ \ \mathtt{ac}(\mathtt{a},\mathtt{b})\ \ \mathtt{if}\ \ \mathtt{s}(\mathtt{a},\mathtt{b}),\ \ \mathtt{a} <> \mathtt{b}.$$

and both $\mathtt{s}(\mathtt{a},\mathtt{b})$ and $\mathtt{a} <> \mathtt{b}$ are contained in $s_0 \cup M$. Thus, we can easily verify that $t = \langle s_0, A, s_1 \rangle$, where $A = \{\mathtt{ac}(\mathtt{a},\mathtt{b})\}$ and $s_1 = \{\mathtt{f}(\mathtt{a})\}$ is a legal state transition: $PD^t$ contains a single causation rule

$$r_1' :\ \ \mathtt{f}(\mathtt{a})\ \ \mathtt{if}\ \ \mathtt{s}(\mathtt{a},\mathtt{a})\ \ \mathtt{after}\ \ \mathtt{ac}(\mathtt{a},\mathtt{b}).$$

which results from $r_1$ for $\theta = \{\mathtt{X} = \mathtt{a},\ \mathtt{Y} = \mathtt{b}\}$. Clearly, $s_1$ satisfies this rule, as $\mathtt{h}(r'_1) \subseteq s_1$, and $s_1$ is smallest, since $\mathtt{s}(\mathtt{a},\mathtt{a}) \in M$ and $\mathtt{ac}(\mathtt{a},\mathtt{b}) \in A$ holds. On the other hand, $t = \langle s_0, A', s_1\rangle$, where $A' = \{\mathtt{ac}(\mathtt{a},\mathtt{b}),\ \mathtt{ac}(\mathtt{b},\mathtt{b})\}$ is not a legal transition: while $\mathtt{ac}(\mathtt{b},\mathtt{b})$ is a legal action instance, there is no executability condition for it in $PD{\downarrow}^t$, and thus $\mathtt{ac}(\mathtt{b},\mathtt{b})$ is not executable in $PD$ w.r.t. $s_0$.

### 2.2.3 Plans

After having defined state transitions, we now formalize plans as suitable sequences of states transitions which lead from an initial state to some success state which satisfies a given goal.

**Definition 2.14** A sequence of state transitions $T = \langle\langle s_0, A_1, s_1\rangle, \langle s_1, A_2, s_2\rangle, \ldots, \langle s_{n-1}, A_n, s_n\rangle\rangle$, $n \geq 0$, is a *trajectory* for $PD$, if $s_0$ is a legal initial state of $PD$ and all $\langle s_{i-1}, A_i, s_i\rangle$, $1 \leq i \leq n$, are legal state transitions of $PD$.

Note that in particular, $T = \langle\rangle$ is empty if $n = 0$.

**Definition 2.15** Given a planning problem $\mathcal{P} = \langle PD, q\rangle$, where $q$ has form (4), a sequence of action sets $\langle A_1, \ldots, A_i\rangle$, $i \geq 0$, is an *optimistic plan* for $\mathcal{P}$, if a trajectory $T = \langle\langle s_0, A_1, s_1\rangle, \langle s_1, A_2, s_2\rangle, \ldots, \langle s_{i-1}, A_i, s_i\rangle\rangle$ in $PD$ exists such that $T$ establishes the goal, i.e., $\{g_1, \ldots g_m\} \subseteq s_i$ and $\{g_{m+1}, \ldots, g_n\} \cap s_i = \emptyset$.

The notion of optimistic plan amounts to what in the literature is defined as "plan" or "valid plan" etc. The term "optimistic" should stress the credulous view underlying this definition, with respect to planning domains that provide only incomplete information about the initial state of affairs and/or bear nondeterminism in the action effects, i.e., alternative state transitions.

In such domains, the execution of an optimistic plan $P$ is not a guarantee that the goal will be reached. We therefore resort to secure plans (alias conformant plans), which are defined as follows.

**Definition 2.16** An optimistic plan $\langle A_1, \ldots, A_n\rangle$ is a *secure plan*, if for every legal initial state $s_0$ and trajectory $T = \langle\langle s_0, A_1, s_1\rangle, \ldots, \langle s_{j-1}, A_j, s_j\rangle\rangle$ such that $0 \leq j \leq n$, it holds that (i) if $j = n$ then $T$ establishes the goal, and (ii) if $j < n$, then $A_{j+1}$ is executable in $s_j$ w.r.t. $PD$, i.e., some legal transition $\langle s_j, A_{j+1}, s_{j+1}\rangle$ exists.

Observe that plans admit in general the concurrent execution of actions at the same time. However, in many cases the concurrent execution of actions may not be desired (and explicitly prohibited, as discussed below), and attention focused to plans with one action at a time. More formally, we call a plan $\langle A_1, \ldots, A_n\rangle$ *sequential* (or *non-concurrent*), if $|A_j| \leq 1$, for all $1 \leq j \leq n$.

## 2.3 Enhanced Syntax

While the language presented in Section 2.1 is complete and allows for a succinct semantics definition, it can be enhanced w.r.t. user-friendliness. E.g. it is inconvenient to write `initially` in front of each initial state constraint, having an `initially` section in which each rule is interpreted as an initial state constraint would be more desirable. In addition, some frequently occurring patterns can be identified for which macros will be defined for convenience and readability.

### 2.3.1 Partitions

The specification of a planning domain $PD = \langle \Pi, \langle D, R\rangle\rangle$ (respectively, of a planning problem $\mathcal{P} = \langle\langle \Pi, \langle D, R\rangle\rangle, q\rangle$) can be seen as being partitioned into

- the background knowledge $\Pi$
- $F_D$, the fluent declarations in $D$
- $A_D$, the action declarations in $D$
- $I_R$, the initial state constraints in $R$
- $C_R$, the causation rules and executability conditions in $R$
- the query (or goal) $q$.

In the sequel, we will denote a planning problem as follows:

```
fluents :    F_D
actions :    A_D
always :     C_R
initially :  I_R
goal :       q
```

where each construct in $F_D$, $A_D$, $C_R$, and $I_R$ is terminated by ".". The background knowledge is assumed to be represented separately.

### 2.3.2 Macros

In the following, we will define several macros which allow for a concise representation of frequently used concepts. Let $\mathtt{a} \in \mathcal{L}^+_{act}$ denote an action atom, $\mathtt{f} \in \mathcal{L}_{fl}$ a fluent literal, B a (possibly empty) sequence $b_1, \ldots, b_k$, not $b_{k+1}, \ldots$, not $b_l$ where each $b_i \in \mathcal{L}_{fl,typ}$, $i = 1, \ldots, l$, and A a (possibly empty) sequence $a_1, \ldots, a_m$, not $a_{m+1}, \ldots$, not $a_n$ where each $a_j \in \mathcal{L}$, $j = 1, \ldots, n$.

**Inertia** In planning it is often useful to declare some fluents as inertial, which means that these fluents keep their truth values in a state transition, unless explicitly affected by an action. In the AI literature this has been studied intensively and is referred to as the *frame problem* [51, 62].

To allow for an easy representation of this kind of situation, we have enhanced the language by the shortcut

$$\texttt{inertial f if B after A.} \quad \Leftrightarrow \quad \texttt{caused f if not} \neg\texttt{.f, B after f, A.}$$

**Defaults** A default value of a fluent in the planning domain can be expressed by the shortcut

$$\texttt{default f.} \quad \Leftrightarrow \quad \texttt{caused f if not} \neg\texttt{.f.}$$

This default is in effect unless there is evidence to the opposite value of fluent f, given through some other causation rule.

**Totality** For reasoning under incomplete, but total knowledge we introduce

$$\texttt{total f if B after A.} \quad \Leftrightarrow \quad \begin{array}{l}\texttt{caused f if not } -\texttt{f, B after A.}\\ \texttt{caused } -\texttt{f if not f, B after A.}\end{array}$$

where f must be positive.

**State Integrity** It is very common to formulate integrity constraints for states (possibly referring to the respective preceding state). To this end, we define the macro

$$\texttt{forbidden B after A} \quad \Leftrightarrow \quad \texttt{caused false if B after A}$$

**Nonexecutability** Sometimes it is more intuitive to specify when some action is not executable, rather than when it is. To this end, we introduce

$$\texttt{nonexecutable a if B} \quad \Leftrightarrow \quad \texttt{caused false after a, B}$$

Note that because of this definition, `nonexecutable` is stronger than `executable`, so in case of conflicts, `executable` is overridden by `nonexecutable`.

**Non-concurrent Plans** Finally, `noConcurrency` disallows the simultaneous execution of actions. We define

$$\texttt{noConcurrency} \quad \Leftrightarrow \quad \texttt{caused false after } \texttt{a}_1, \texttt{a}_2.$$

where $\texttt{a}_1$ and $\texttt{a}_2$ range over all possible actions such that $\texttt{a}_1, \texttt{a}_2 \in \mathcal{L}_{PD} \cap \mathcal{L}_{act}$ and $\texttt{a}_1 \neq \texttt{a}_2$.

In all macros, "`if B`" (resp., "`after A`") can be omitted, if B (resp. A) is empty. We reserve the possibility of including further macros in future versions of $\mathcal{K}$.

# 3 Knowledge Representation in $\mathcal{K}$

In this section, the use of $\mathcal{K}$ for modeling planning problems is explored by examples. Special attention is given to features and techniques which distinguish $\mathcal{K}$ from similar languages.

## 3.1 Deterministic Planning with Complete Knowledge

We first study a simple setting in which the planning domain is not subject to nondeterminism and the planning agent has complete knowledge of the state of affairs. For later reference, we formally introduce the following notion.

**Definition 3.1** Let $PD$ be a planning domain. Then, a legal transition $\langle s, A, s_1\rangle$ in $PD$ is *determined*, if $s_1 = s_2$ holds for every possible legal transition $\langle s, A, s_2\rangle$ (i.e., executing $A$ on $s$ leads to a unique new state). We call $PD$*deterministic*, if all legal transitions in it are determined.

Consider first the planning problem depicted in Figure 1, which is set in the blocksworld. This problem illustrates the famous Sussman anomaly [66].

We will first describe the planning domain $PD_{bwd} = \langle \Pi_{bw}, \langle D_{bwd}, R_{bwd}\rangle\rangle$ of blocksworld. It involves distinguishable blocks and a table. Blocks and the table can serve as locations on which other blocks can be put (a block can hold at most one other block, while the table can hold arbitrarily many blocks). We thus define the notions of `block` and `location` in the background knowledge $\Pi_{bw}$ as follows:

```
block(a). block(b). block(c).
location(table).
location(B) :- block(B).
```

For representing states, we declare two fluents in $F_{D_{bwd}}$: `on` states the fact that some block resides on some location, `occupied` is true for a location, if its capacity of holding blocks is exhausted.

```
fluents :   on(B,L) requires block(B), location(L).
            occupied(B) requires location(B).
```

Only one action is declared in $A_{D_{bwd}}$: `move` represents moving a block to some location (implicitly removing it from its previous location).

```
actions :   move(B,L) requires block(B), location(L).
```

Let us now specificy the initial state constraints $I_{R_{bwd}}$. For the initial state, `occupied` does not have to be specified, as it follows from knowledge about `on`. Note that only positive facts are stated for `on`, nevertheless the initial state is unique because the fluent `on` is interpreted under the closed world assumption (CWA) [59], i.e. if `on(B,L)` does not hold, we assume that it is false.

```
initially : on(a,table). on(b,table). on(c,a).
```

Next, we specify causation rules and executability conditions $C_{R_{bwd}}$. First a static rule is given, defining `occupied` for blocks if some other block is `on` them.

```
always :    caused occupied(B) if on(B1,B), block(B).
```

A `move` action is executable if the block to be moved and the target location are distinct (a block cannot be moved onto itself). A move is not executable if either the block or the target location is occupied.

```
            executable move(B,L) if B <> L.
            nonexecutable move(B,L) if occupied(B).
            nonexecutable move(B,L) if occupied(L).
```

The action effects are defined by dynamic rules. They state that a moved block is on the target location after the move, and that a block is not on the location on which it resided before it was moved.

```
            caused on(B,L) after move(B,L).
            caused -on(B,L1) after move(B,L), on(B,L1), L <> L1.
```

Next we state that the fluent `on` should stay true, unless it becomes false explicitly. Note that we need not specify this property for `occupied`, as it follows from `on` via the static rule.

```
            inertial on(B,L).
```

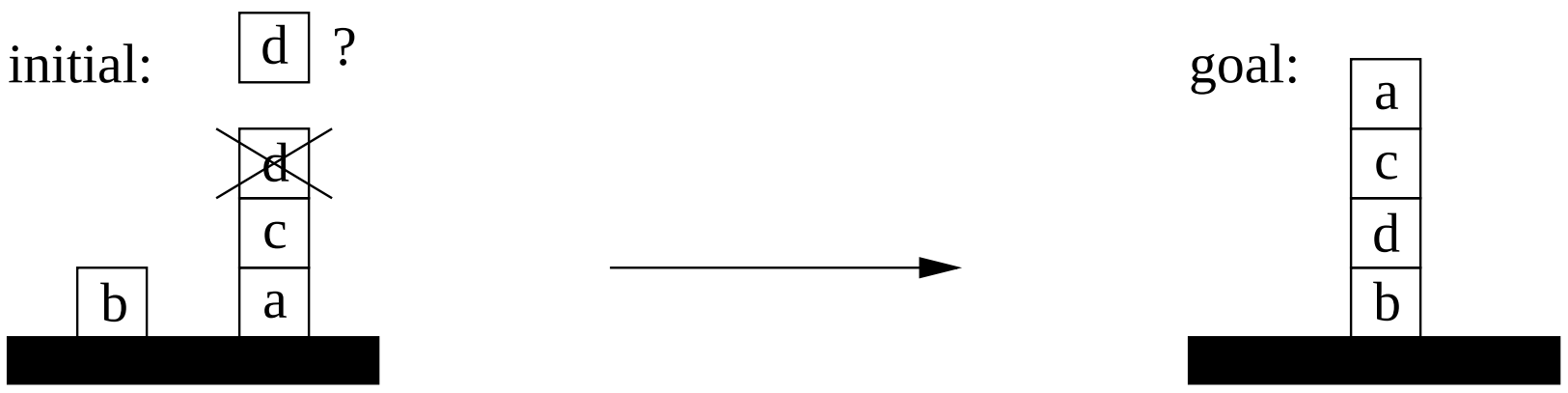

Figure 2: A Blocks World example with incomplete initial state.

It is worthwhile noting that in this example the fluents are represented positively. The negation of fluents is usually implicit via the closed world assumption. There is one exception in a rule describing an action effect: Here the negation becomes known explicitly, and its purpose is the termination of the inertial truth of an instance of on.

In order to solve the original planning problem, we associate the following goal $q_{bwd}$ for plan length 3 to $PD_{bwd}$, yielding $\mathcal{P}_{bwd}$:

$$\texttt{goal}: \quad \texttt{on(c,b)},\ \texttt{on(b,a)},\ \texttt{on(a,table)}\ ?\ (3)$$

$\mathcal{P}_{bwd}$ allows a single sequential plan of length 3:

$$\langle\{\texttt{move(c,table)}\},\{\texttt{move(b,a)}\},\{\texttt{move(c,b)}\}\rangle$$

Thus, the above plan requires to first move c on the table, then to move b on top of a, and, finally, to move c on b. It is easy to see that this sequence of actions leads to the desired goal. Since this domain is deterministic and has a unique initial state, all optimistic plans are also secure.

## 3.2 Planning with Incomplete Initial State Descriptions

In the example of section 3.1, it is assumed that the initial state is correct (with respect to the domain in question) and fully specified (thus unique). In this section we explore how these implicit requirements can be weakened.

As an accompanying example problem, suppose that there is a further block d in the original planning problem of Figure 1. The exact location if d is unknown, but we know that it is not on top of c. Furthermore, there is a slightly different goal involving d. The problem is depicted in Figure 2. We will define a corresponding planning domain $PD_{bwi} = \langle \Pi_{bwi}, \langle D_{bwi}, R_{bwi}\rangle\rangle$ by extending $PD_{bwd}$. The additional knowledge about the initial state is represented by adding $-\texttt{on(d,c)}$. to $I_{R_{bwi}}$. and the background knowledge $\Pi_{bwi}$ is obviously enriched by the fact block(d).

Let us first consider the necessary extensions for handling cases in which the initial state description cannot be assumed to be correct (e.g., when completing the partial initial state description, incorrect initial states can arise). The following conditions should be verified for each block: (i) It is on top of a unique location, (ii) it does not have more than one block on top of it, and (iii) it is supported by the table (i.e., it is either on the table or on a stack of blocks which is on the table) [44].

It is straightforward to formulate conditions (i) and (ii) and include them into $I_{R_{bwi}}$:

```
initially: forbidden on(B,L), on(B,L1), L <> L1.
           forbidden on(B1,B), on(B2,B), block(B), B1 <> B2.
```

For condition (iii) we add a fluent supported to $F_{D_{bwi}}$, which should be true for any block in a legal initial state:

$$\texttt{fluents:} \quad \texttt{supported(B) requires block(B).}$$

We add the definition of supported and a constraint stating that each block must be supported to $I_{R_{bwi}}$.

```
initially: caused supported(B) if on(B,table).
           caused supported(B) if on(B,B1), supported(B1).
           forbidden not supported(B).
```

Any planning problem involving the domain defined so far does not admit any plan if the initial state is either incorrectly specified or incomplete in the sense that not all block locations are known (as supported will not hold for these blocks). Note that the action move preserves the properties (i),(ii), (iii) above for sequential plans; it is therefore not necessary to check these properties in all states if concurrent actions are not allowed.

Next we show how incomplete initial states can be completed in $\mathcal{K}$. To this end, we use the keyword total (defined in section 2.3.2), and simply add total on(X, Y). to $I_{R_{bwi}}$. In this way, all possible completions w.r.t. on(X, Y) serve as candidate initial states, only some of which satisfy the initial state constraints, making them legal initial states. E.g. the state in which on(d, a) holds is not legal as the constraint which checks condition (ii) is violated.

Finally, let us consider the planning problem $\mathcal{P}_{bwi} = \langle PD_{bwi}, q_{bwi} \rangle$, where $q_{bwi}$ is

```
goal:      on(a,c), on(c,d), on(d,b), on(b,table) ? (j)
```

Usually, when dealing with incomplete knowledge, we look for plans which establish the goal for any legal initial state (in this particular case case no matter whether on(d, b) or on(d, table) holds), so we are interested in *secure plans*. The following secure sequential plan exists for $\mathcal{P}_{bwi}$ and $j = 4$:

$$\langle \{\texttt{move(d,table)}\}, \{\texttt{move(d,b)}\}, \{\texttt{move(c,d)}\}, \{\texttt{move(a,c)}\} \rangle$$

It is easily verifiable that this plan works on each legal initial state: Since d is not occupied in any legal initial state, the first action can always be executed.

In some cases, one is interested in a plan which works for some possible initial state: For $\mathcal{P}_{bwi}$ an optimistic plan exists for $j = 2$:

$$\langle \{\texttt{move(c,d)}\}, \{\texttt{move(a,c)}\} \rangle$$

It works only for the initial state in which on(d, b) holds, and fails for all other admissible initial states. Hence it is not a secure plan.

### 3.3 Nondeterministic Action Effects

Let us now focus on domains comprising nondeterministic action effects. To this end we will turn our attention to the "bomb in the toilet" problem [52] and its variations. We will describe these domains gradually, starting with two versions which involve deterministic action effects and incomplete initial state specifications, in which the representation techniques from section 3.2 are applied. Only after these, a variant comprising nondeterministic action effects and some additional elaborations are presented. We employ a naming convention which is due to [6].

**BT($p$) - Bomb in toilet with $p$ packages** We have been alarmed that there is a bomb (exactly one) in a lavatory. There are $p$ suspicious packages which could contain the bomb. There is one toilet bowl, and it is possible to dunk a package into it. If the dunked package contained the bomb, the bomb is disarmed.

For the $\mathcal{K}$ encoding, the background knowledge $\Pi_{bt}$ consists of a definition of the packages:

$$\texttt{package}(1).\quad \texttt{package}(2).\quad \ldots\quad \texttt{package}(p).$$

We use two fluents: armed(P) holds if package P contains an armed bomb (this is an inertial property), and unsafe expresses the fact that there are armed bombs. Only one action, dunk(P), is required, which is always executable and the effect of which is that package P is no longer armed.

For the initial state, totalarmed(P). expresses the fact that the armed bomb might be in any package P, while forbiddenarmed(P), armed(P1), P <> P1. enforces that at most one package can contain an armed bomb. The statement forbiddennotunsafe. is included to guarantee that at least one package contains an armed bomb in the initial state.

The goal is to achieve a state in which no armed bomb exists, i.e. which is notunsafe. This goal $q_{bomb}$ will be the same for all following variations of the bomb in toilet problems, the respective plan lengths $j$ will be stated for each problem. We thus arrive at the following planning problem $\mathcal{P}_{bt} = \langle PD_{bt}, q_{bomb}\rangle$:

```
fluents:   armed(P) requires package(P).
           unsafe.
actions:   dunk(P) requires package(P).
always:    inertial armed(P).
           caused -armed(P) after dunk(P).
           caused unsafe if armed(P).
           executable dunk(P).
initially: total armed(P).
           forbidden armed(P), armed(P1), P <> P1.
           forbidden not unsafe.
goal:      not unsafe ? (j)
```

Note that in the formulation of this simple domain there is only one deterministic action, while the initial state is incomplete since it is not known which of the $p$ packages contains the bomb.

Usually, a plan should be produced which establishes the goal no matter in which package the bomb is in, so we look for a secure plan. If concurrent actions are allowed, the following secure plan for $j = 1$ (dunking all packages at the same time) can be found:

$$\langle\{\texttt{dunk}(1), \ldots, \texttt{dunk}(p)\}\rangle$$

A secure sequential plan consists of dunking all packages sequentially, so $j = p$:

$$\langle\{\texttt{dunk}(1)\}, \ldots, \{\texttt{dunk}(p)\}\rangle$$

Any permutation of these action sets is also a valid secure plan.

**BTC($p$) - Bomb in toilet with certain clogging** Let us now consider a slightly more elaborate version of the problem: Assume that dunking a package clogs the toilet, making further dunking impossible. The toilet can be unclogged by flushing it. The toilet is assumed to be unclogged initially. Note that this domain still comprises only deterministic action effects.

We extend $PD_{bt} = \langle \Pi_{bt}, \langle D_{bt}, R_{bt} \rangle \rangle$ to $PD_{btc} = \langle \Pi_{bt}, \langle D_{btc}, R_{btc} \rangle \rangle$ by adding a new fluent, `clogged`, and a new action, `flush`, to $D_{btc}$:

```
fluents :   clogged.
actions :   flush.
```

`clogged` is inertial, is a deterministic effect of `dunk`, and is terminated by `flush`. `flush` is always executable, so the following rules are added to $C_{R_{btc}}$:

```
always :    inertial clogged.
            caused −clogged after flush.
            caused clogged after dunk(P).
            executable flush.
```

The executability statement for `dunk` has to be modified, as `dunk` is not executable if the toilet is clogged.

```
            executable dunk(P) if not clogged.
```

Since `clogged` is assumed not to hold initially, and since it is interpreted under the CWA, nothing has to be added to $I_{R_{btc}}$.

For the planning problem $\mathcal{P}_{btc} = \langle PD_{btc}, q_{bomb} \rangle$ we are only interested in sequential plans, as dunking and flushing concurrently is not permitted. A minimal secure plan can be found for $j = 2p - 1$:

$$\langle \{\mathtt{dunk(1)}\}, \{\mathtt{flush}\}, \{\mathtt{dunk(2)}\}, \ldots, \{\mathtt{flush}\}, \{\mathtt{dunk(p)}\} \rangle$$

Again, the action sets containing `dunk` actions can be arbitrarily permuted, as long as the `flush` actions are executed between the `dunk` actions.

**BTUC($p$) - Bomb in toilet with uncertain clogging** Consider a further elaboration of the domain, in which `clogged` may or may not be an effect of dunking. In other words, the action `dunk` has a nondeterministic effect, and the toilet is clogged or not clogged after having executed `dunk`.

This behavior is modeled by declaring `clogged` to be `total` after `dunk` has occurred. Therefore the action effect

```
            caused clogged after dunk(P).
```

in $PD_{btc}$ is modified to

```
            total clogged after dunk(P).
```

yielding the planning domain $PD_{btuc}$. The planning problem $\mathcal{P}_{btuc} = \langle PD_{btuc}, q_{bomb} \rangle$ admits the same secure plans as $\mathcal{P}_{btc}$.

**BMTC($p$,$t$), BMTUC($p$,$t$) - Bomb in toilet with multiple toilets** Yet another elaboration is to assume that several toilet bowls ($t$, rather than just one) are available in the lavatory. The modifications to $PD_{btc}$ yielding $PD_{bmtc} = \langle \Pi_{bmt}, \langle D_{bmtc}, R_{bmtc} \rangle \rangle$ and to $PD_{btuc}$ yielding $PD_{bmtuc} = \langle \Pi_{bmt}, \langle D_{bmtuc}, R_{bmtuc} \rangle \rangle$ are rather straightforward.

The background knowledge $\Pi_{bt}$ is simply extended to contain also a definition of the $t$ toilets, by adding:

```
toilet(1).  toilet(2).  ...  toilet(t).
```

arriving at $\Pi_{bmt}$. The fluent and action declarations for `clogged`, `dunk`, and `flush` must be parametrised w.r.t. the affected toilet. The updated definitions w.r.t. $D_{btc}$ (resp. $D_{btuc}$) are as follows:

```
clogged(T) requires toilet(T).
dunk(P,T) requires package(P), toilet(T).
flush(T) requires toilet(T).
```

Furthermore, each occurrence of `clogged`, `dunk`, and `flush` in $R_{btc}$ (resp. $R_{btuc}$) must be updated by adding a variable `T` (representing the toilet) to its parameters.

Since multiple resources can be used concurrently here, we add some concurrency conditions for the actions to $PD_{btc}$ (resp. $PD_{btuc}$): `dunk` and `flush` should never be executed concurrently on any toilet. Furthermore, at most one package should be dunked into a toilet, and any package should be dunked in at most one toilet at a time. These conditions are captured by the following rules:

```
always:      nonexecutable dunk(P,T) if flush(T).
             nonexecutable dunk(P,T) if dunk(P1,T), P <> P1.
             nonexecutable dunk(P,T) if dunk(P,T1), T <> T1.
```

In total, $\langle D_{bmtuc}, R_{bmtuc} \rangle$ of $PD_{bmtuc}$ looks as follows:

```
fluents:     clogged(T) requires toilet(T).
             armed(P) requires package(P).
             unsafe.
actions:     dunk(P,T) requires package(P), toilet(T).
             flush(T) requires toilet(T).
always:      inertial armed(P).
             inertial clogged(T).
             caused -clogged(T) after flush(T).
             caused -armed(P) after dunk(P,T).
             total clogged(T) after dunk(P,T).
             caused unsafe if armed(P).
             executable flush(T).
             executable dunk(P,T) if not clogged(T).
             nonexecutable dunk(P,T) if flush(T).
             nonexecutable dunk(P,T) if dunk(P1,T), P <> P1.
             nonexecutable dunk(P,T) if dunk(P,T1), T <> T1.
initially:   total armed(P).
             forbidden armed(P), armed(P1), P <> P1.
             forbidden not unsafe.
```

The secure plans for $\mathcal{P}_{bmtc} = \langle PD_{bmtc}, q_{bomb} \rangle$ and $\mathcal{P}_{bmtuc} = \langle PD_{bmtuc}, q_{bomb} \rangle$ are similar to those for $\mathcal{P}_{btc}$ and $\mathcal{P}_{btuc}$, respectively. The differences are that up to $t$ `dunk` and `flush` actions, respectively, can be executed in parallel (so the plans are no longer sequential), and that $t-1$ flushing actions can be saved since no final flushing is required for any toilet. Therefore any secure plan consists of $2p - t$ actions and in $q_{bomb}$, the minimal plan length is: $j = 2\lceil \frac{p}{t} \rceil - 1$.

### 3.4 Knowledge Based Encoding of Nondeterministic Action Effects

In this section, alternative planning domains for bomb in toilet will be presented. These encodings will be based on states of knowledge, a distinguishing feature of $\mathcal{K}$, rather than states of the world as in the previous sections. We will use the same background knowledge $\Pi_{bt}$ (resp. $\Pi_{bmt}$) and the same goal $q_{bomb}$ with the same values for the plan length $j$ as in section 3.3.

**BT($p$)** In section 3.3 we have represented the initial situation by means of totalization on `armed(P)`, leading to multiple initial states, corresponding to different possible states of the world. From the knowledge perspective, nothing is known about `armed(P)` (and `-armed(P)`), so the initial situation can be represented by one state in which neither `armed(P)` nor `-armed(P)` holds. The action `dunk(P)` has the effect that -armed(P) is known to hold, and `-armed(P)` is inertial. We state the planning domain $PD_{btk}$ as follows:

```
fluents:   armed(P) requires package(P).
           unsafe.
actions:   dunk(P) requires package(P).
always:    inertial -armed(P).
           caused -armed(P) after dunk(P).
           caused unsafe if not -armed(P).
           executable dunk(P).
```

The advantage of this encoding is that multiple initial states do not have to be dealt with. Note that in this formulation it does not make sense to encode the restriction that exactly one package is armed, as nothing is known about the armed status whatsoever, so reasoning about what conditions this knowledge should comply with, if we had it, is superfluous. Furthermore, since the domain is deterministic, optimistic and secure plans coincide.

**BTC($p$)** $PD_{btck}$ is extended from $PD_{btk}$ in the same way as $PD_{btc}$ was obtained from $PD_{bt}$ in section 3.3, i.e. by adding declarations for `clogged` and `flush`, adding rules for action effects w.r.t. `clogged`, defining `clogged` to be inertial, stating `flush` to be always executable, and by modifying the executability condition for `dunk(P)`.

Note that in this encoding `clogged` is still interpreted under the CWA.

**BTUC($p$)** In the variant with uncertain clogging, the effect of `dunk(P)` is that the truth of `clogged` is unknown. $\mathcal{K}$ has the capability of representing a state in which neither `clogged` nor `-clogged` holds, but to do this, we should no longer interpret `clogged` under the CWA, as we would not like to assume that `clogged` does not hold if it is unknown. For this reason `inertial -clogged.` is included, and for the initial state, it must be stated explicitly that the toilet is unclogged.

Unfortunately, there is no construct in $\mathcal{K}$, with which an action effect of some fluent being unknown can be expressed directly. However, it is possible to modify the inertial rules for `clogged` and `-clogged`, so that inertia applies only if no package has been dunked. That means that dunking stops inertia for `clogged`, and `clogged` will be unknown unless it becomes known otherwise. Since this technique encodes inertia under some conditions, we call it *conditional inertia*.

To achieve this, a new fluent `dunked` is introduced, which holds immediately after `dunk(P)` occurred for some package `P`. The `inertial` macros are then extended by the additional condition. The precise meaning of the resulting program is that neither `clogged` nor `-clogged` will hold after `dunk(P)` has been executed for some package `P`, unless one of them is caused by some other rule different from inertia.

To summarize, the following is added to $PD_{btck}$:

```
fluents :    dunked.
always :     inertial clogged if not dunked.
             inertial -clogged if not dunked.
             caused dunked after dunk(P).
             caused -clogged after flush.
             executable dunk(P) if -clogged.
initially :  -clogged.
```

while a few statements are dropped:

```
always :     inertial clogged.
             caused clogged after dunk(P).
             executable dunk(P) if not clogged.
```

yielding $PD_{btuck}$.

Note that also $PD_{btuck}$ is deterministic and has a unique initial state, so optimistic and secure plans coincide. This example shows that it is possible to find an encoding which requires a substantially less complex solver by using techniques, which exploit the "state of knowledge" paradigm of the language $\mathcal{K}$. We would like to point out that this is not a contradiction to complexity results in section 4 below (finding secure plans is more complex than finding optimistic plans): BTUC($p$) per se is an easy problem (it is solvable in linear time), it is just the representation requiring examination of alternatives, which made it look hard.

**BMTC($p$,$t$), BMTUC($p$,$t$)** As in section 3.3, a generalization to domains involving multiple toilets is straightforward and can be achieved by applying the changes described there, resulting in the planning domains $PD_{bmtck}$ and $PD_{bmtuck}$, respectively. Find $PD_{bmtuck}$ as an example below ($\Pi_{bmt}$ is omitted):

```
fluents :    clogged(T) requires toilet(T).
             armed(P) requires package(P).
             dunked(T) requires toilet(T).
             unsafe.
actions :    dunk(P, T) requires package(P), toilet(T).
             flush(T) requires toilet(T).
always :     inertial -armed(P).
```

```
                inertial clogged(T) if not dunked(T).
                inertial -clogged(T) if not dunked(T).
                caused dunked(T) after dunk(P,T).
                caused -clogged(T) after flush(T).
                caused -armed(P) after dunk(P,T).
                caused unsafe if not -armed(P).
                executable flush(T).
                executable dunk(P,T) if -clogged(T).
                nonexecutable dunk(P,T) if flush(T).
                nonexecutable dunk(P,T) if dunk(P1,T), P <> P1.
                nonexecutable dunk(P,T) if dunk(P,T1), T <> T1.
     initially: -clogged(T).
```

Also in this case the resulting problem domains are deterministic and hence optimistic plans and secure plans coincide. This indicates that planning problems of this section can be solved faster than those of section 3.3. Indeed, we have observed this also experimentally [12]; the encodings of section 3.4 can often be solved several orders of magnitudes faster than those of section 3.3 in the $\texttt{DLV}^{\mathcal{K}}$ system prototype.

### 3.5 Discussion

As we have seen in the preceding subsections, the use of knowledge states instead of world states allows us to represent planning scenarios in which certain information remains open, or is (deliberatively) dropped under the proviso that it is not relevant to the planning problems that are considered. However, the `total` primitive provides a simple means to switch from knowledge states to world states, and thus our approach fully supports conventional world state planning.

An important advantage which our language offers is that it also enables planning where world states are projected to a subset of fluents of interest, leaving the details of other fluents open. It thus supports to some extent *focusing* in the problem representation, by restricting attention to those fluents whose value may have an influence on the evolution of the world depending on the actions that are taken.

For example, if the toilets in the bomb in the toilet domain would be colored, and an action `paint(T,C)` would be available which causes the color of toilet `B` to become `C`, represented by the fluent `color(T,C)`, then the fluent `color` is not relevant for the planning problems considered in Sections 3.3 and 3.4. Thus, the value of this fluent may be left open, and no totalization statement on `color` is needed on the problem representation.

The question then is how relevance can be (efficiently) determined and exploited by the user. In general, efficient automatic support will be difficult to achieve, since it requires analysis of the planning domain which involves intractable computational problems. However, using adapted results about relevance in logic programming, cf. [9], under some assertions syntactic criteria may be used to exclude (part of the) fluents which are irrelevant for a goal. In the above example, given a natural representation we would find out that `color(T,C)` is not relevant for `unsafe`. Sophisticated usage of `total` remains with the user at the moment, and developing automated support is an interesting research topic.

Another issue concerns the use of knowledge states versus world states, even with respect to fluents that are relevant for achieving the planning goal. Here, we must take into account the underlying assumption of taking actions depending on a state of knowledge (where in case of incomplete information, default assumptions might be used) or a state of affairs, respectively.

For example, if a robot is in front of a door, and wants to pass through it, it needs to know whether the door is open or not. In our approach, we may represent this by the following statements:

$r_1$ : `-open if not open after check_door.`
$r_2$ : `open if not -open after check_door.`
$e$ : `executable check_door if not open, not -open.`

That is, after checking the state of the door, we know whether it is open or not (both is possible), and a secure plan must handle both cases appropriately. The `check_door` action is only executable if the state is not known yet – otherwise doing it would be superfluous, assuming that the robot's state correctly models the world. Thus, under knowledge state planning, a global plan may naturally include the action `check_door`, assuming that its status is unknown in the current state. However, under world-state planning, such an action would always be superfluous as the value of `open` is known. Accordingly, if we add the statement `total open.`, then a plan including `check_door` is no longer feasible; this, however, is not a flaw, since it simply reflects that the precondition for executing the sensing action, namely that the door status is unknown, does never apply. In the same line, we can find examples where adding total statements render secure plans insecure, or where new optimistic and secure plans emerge. On the other hand, by forgetting the status of fluents, we might find plans for problems where world-state planning has no plan.

We may explain these observations by reminding that knowledge state planning, in our approach, is planning under (default) assumptions made on incomplete information, which are represented in the planning domain by the use of default literals and select one of the two possible values of a fluent. These assumptions may turn out inappropriate in reality, and a plan may become infeasible. Security of a plan is relative to the emerging states of knowledge and the assumptions that were made in selecting the actions. This looks refutable, but a moment of reflection should convince that this incorporates *qualitative decision making* in terms of default principles into the planning process. Any statement `total f.` is an unconditional *implicit sensing action*, which refines the knowledge state by reporting the status of the fluent in the new state.

We thus may proceed in planning as follows: try to find an optimistic or secure plan, and then evaluate feasibility of the plan under refined knowledge states, by adding suitable `total` statements. Here, not necessarily all fluents have to be totalized, but merely the relevant ones. In case no plan exists, a refinement of the knowledge states may be attempted at the initial state. In particular, if incompleteness is just given in the initial state, but each fluent is, by the causal rules, defined in each future state, then one should describe the properties known to hold in the beginning, totalize the (relevant) fluents of the initial state, and ask for a secure plan (cf. section 3.2, the interested reader is encouraged to identify the relevant instances of `on(X, Y)` for totalization w.r.t. the goal there). Exploring the use of totalization, and developing a methodology for this process is an interesting issue for further work.

## 4 Complexity of $\mathcal{K}$

We now turn to the computational complexity of planning in our language $\mathcal{K}$. In this section, we present the results of a detailed study of major planning issues in the propositional case. Results for the case of general planning problems (with variables) may be obtained by applying suitable complexity upgrading techniques (cf. [30]). We call a planning domain $PD$ (resp., planning problem $\mathcal{P}$) *propositional*, if all predicates in it have arity 0, and thus it contains no variables.

### 4.1 Main Problems Studied

In our analysis, we consider the following three problems:

**Optimistic Planning** Decide, given a propositional planning problem $\langle PD, q\rangle$, whether some optimistic plan exists.

**Security Checking** Decide, given an optimistic plan $P = \langle A_1, \ldots, A_n\rangle$ for a propositional planning problem $\langle PD, q\rangle$, whether $P$ is secure.

**Secure Planning** Decide, given a propositional planning problem $\langle PD, q\rangle$, whether some secure plan exists.

We remark here that the formulation of security checking is, strictly speaking, a *promise problem*, since it is *asserted* that $P$ is an optimistic plan, which can not be checked in polynomial time in general (and thus legal inputs can not be recognized easily). However, the complexity results that we derive below would remain the same, even if $P$ were not known to be an optimistic plan.

We assume that the reader has some knowledge of basic concepts of computational complexity theory; see [54, 7] for a background and further sources. In particular, we assume familiarity with the well-known complexity classes P, NP, co-NP, and PSPACE. The classes $\Sigma_k^P$ (resp., $\Pi_k^P$), $k \geq 0$ of the Polynomial Hierarchy PH = $\bigcup_{k\geq 0} \Sigma_k^P$ are defined by $\Sigma_0^P = \Pi_0^P = \mathrm{P}$ and $\Sigma_k^P = \mathrm{NP}^{\Sigma_{k-1}^P}$ (resp., $\Pi_k^P$ = co-$\Sigma_k^P$), for $k \geq 1$. The latter model nondeterministic polynomial-time computation with an oracle for problems in $\Sigma_{k-1}^P$. Furthermore, $\mathrm{D}^P = \{L \cap L' \mid L \in \mathrm{NP}, L' \in \text{co-NP}\}$ is the logical “conjunction” of NP and co-NP, and NEXPTIME is the class of problems decidable by nondeterministic Turing machines in exponential time. We recall that $\mathrm{NP} \subseteq \mathrm{D}^P \subseteq$ PH $\subseteq$ PSPACE=NPSPACE $\subseteq$ NEXPTIME holds, where NPSPACE is the nondeterministic analog of PSPACE. It is generally believed that these inclusions are strict, and that PH is a true hierarchy of problems with increasing difficulty. Note that NEXPTIME-complete problems are *provably intractable*, i.e., exponential lower bounds can be proved, while no such proofs for problems in PH or PSPACE are known today.

### 4.2 Overview of Results

We will consider the three problems from above under the following two restrictions:

**1. General vs. proper planning domains** Because of their underlying stable semantics, which is well-known intractable [46], causation rules in domain descriptions can express computationally intractable relationships between fluents. In fact, determining whether for a state $s$ and a set of executable actions $A$ in $s$ some legal transition $\langle s, A, s'\rangle$ to any successor state $s'$ exists in a planning domain $PD$ is intractable in general, since it comprises checking whether a logic program has an answer set. For this reason, we pay special attention to the following subclass of planning domains.

**Definition 4.1** We call a planning domain $PD$ *proper* if, given any state $s$ and any set of actions $A$, deciding whether some legal state transition $\langle s, A, s'\rangle$ exists is polynomial. A planning problem $\langle PD, q\rangle$ is *proper*, if $PD$ is proper.

Proper planning domains are not plagued with intractability of deciding whether doing some actions will violate the dynamic domain axioms, even if they possibly have nondeterministic effects. In fact, we expect that in many scenarios, the domain is represented in a way such that if a set of actions qualifies for execution in a state, then performing these actions is guaranteed to reach a successor state. In such cases, the planning domain is trivially proper. This applies, for example, to the standard STRIPS formalism [20] and many of its variants.

Unfortunately, deciding whether a given planning domain is proper is intractable in general; we thus need syntactic restrictions for taking advantage of this (semantic) property in practice. For obtaining significant lower complexity bounds, we consider in our analysis a very simple class of proper planning domains.

**Definition 4.2** We call a planning domain $PD = \langle \Pi, AD \rangle$ *plain*, if the background knowledge $\Pi$ is empty, and $AD$ satisfies the following conditions:

1. Executability conditions refer only to fluents.
2. No default negation –neither explicit nor implicit through language extensions (such as inertia rules)– is used in the $post$-part of causation rules in the "`always`" section.
3. Given that $\alpha_1, \ldots, \alpha_m$ are all ground actions, $AD$ contains the rules

   `nonexecutable` $\alpha_i$ `if` $\alpha_j$. $\quad 1 \leq i < j \leq m$
   `caused false after not` $\alpha_1$, `not` $\alpha_2$, ... , `not` $\alpha_m$.

We call a planning problem $\mathcal{P} = \langle PD, q \rangle$ *plain*, if $PD$ is plain.

The conditions ensure that every legal state transition $t = \langle s, A, s' \rangle$ must satisfy $|A| = 1$. Thus all optimistic and secure plans must be sequential.

As easily seen, in plain planning domains (which can be efficiently recognized), deciding whether for a state $s$ and an action set $A$ some legal state transition $t = \langle s, A, s' \rangle$ exists is polynomial, since this reduces to evaluating a `not` -free logic program with constraints. Thus, plain planning domains are proper. Furthermore, each legal state transition $t$ in a plain planning domain $PD$ is clearly determined, and thus $PD$ is also deterministic. As discussed below, for many problems plain planning domains harbor already the full complexity of proper planning domains.

We remark that further, more expressive syntactic fragments of proper planning domains can be obtained by exploiting known results on logic programs which are guaranteed to have answer sets, such as stratified logic programs, or order-consistent and odd-cycle free logic programs [17, 10]; the latter allow for expressing nondeterministic action effects. In particular, these results may be applied on the rules obtained from the dynamic causation rules by stripping off their $pre$-parts. We do not investigate this issue further here; stratified planning domains are addressed in [57].

**2. Fixed vs. arbitrary plan length** We analyze the impact of fixing the length $i$ in the query $q = Goal\ ?\ (i)$ of $\langle PD, q \rangle$ to a constant. In general, the length of an optimistic plan for $\langle PD, q \rangle$ can be exponential in the size of the string representing the number $i$ (which, as usual, is represented in binary notation), and even exponential in the size of the string representing the whole input $\langle PD, q \rangle$. Indeed, it may be necessary to pass through an exponential number of different states until a state satisfying the goal is reached. For example, the initial state $s_0$ may describe the value $(0, \ldots, 0)$ of an $n$-bit counter, and the goal description might state that the counter has value $(1, \ldots, 1)$. Assuming an action repertoire which allows, in each state, to increment the value of the counter by 1, the shortest optimistic plan for this problems has $2^n - 1$ steps. (We leave the formalization of this problem in $\mathcal{K}$ as an illustrative exercise to the reader.) This shows that storing a complete optimistic plan in working memory requires exponential space in general. If $i$ is fixed, however, then the representation size of the plan is linear in the size of $\langle PD, q\rangle$.

| planning domain $PD$ | plan length $i$ in query $q = Goal\ ?\ (i)$ | |
|---|---|---|
| | fixed (=constant) | arbitrary |
| general | NP / $\Pi_2^P$ / $\Sigma_3^P$ -complete | PSPACE / $\Pi_2^P$ / NEXPTIME -complete |
| proper | NP / co-NP / $\Sigma_2^P$ -complete | PSPACE / co-NP / NEXPTIME -complete |

Table 1: Complexity Results for Optimistic Planning / Security Checking / Secure Planning (Propositional Case)

**Main complexity results** Our main results on the complexity of $\mathcal{K}$ are compactly summarized in Table 1, and can be explained as follows.

- As for Optimistic Planning, we can avoid exponential space for storing an optimistic plan $P = \langle A_1, \ldots, A_n\rangle$ by generating it *step by step*: we guess a legal initial state $s_0$, and subsequently, one by one, the legal transitions $\langle s_{i-1}, A_i, s_i\rangle$. Since storing one legal transition requires only polynomial workspace and NPSPACE=PSPACE, Optimistic Planning is in PSPACE. On the other hand, propositional STRIPS, which is PSPACE-complete [3], can be easily reduced to planning in $\mathcal{K}$, where the resulting planning problem is plain and thus proper. For fixed plan length, the *whole* optimistic plan has linear size, and thus can be guessed and verified in polynomial time.

- In Security Checking, the optimistic plan $P = \langle A_0, \ldots, A_n\rangle$ to be checked is part of the input, so the binary representation of the plan length is not an issue here. If $P$ is not secure, there must be a legal initial state $s_0$ and a trajectory executing the actions in $A_0, \ldots, A_i$ such that either the execution is stuck, i.e., no successor state $s_i$ exists or the actions in $A_i$ are not executable in $s_i$, or the goal is not fulfilled in the final state $s_n$. Such a trajectory can be guessed and verified in polynomial time with the help of an NP oracle; this places the problem in $\Pi_2^P$. The NP oracle is needed to cover the case where no successor state $s_i$ exists, which reduces to checking whether a logic program has no answer set. In proper planning domains, existence of $s_i$ can be decided in polynomial time, which makes the use of an NP oracle obsolete and lowers the overall complexity from $\Pi_2^P = \text{co-NP}^{\text{NP}}$ to co-NP.

- In Secure Planning, the existence of a secure plan can be decided by composing algorithms for constructing optimistic plans and for security checking. Our membership proofs for deciding the existence of an optimistic plan actually (nondeterministically) construct such a plan, and thus we easily obtain upper bounds on the complexity of Secure Planning from the complexity of the combined algorithm, by using the security check as an oracle. In the case of arbitrary plan length, the use of a $\Pi_2^P$ oracle can be eliminated by a more clever procedure, in which plan security is checked by inspecting all states reachable after $0, 1, 2, \ldots$ steps of the plan. Even if their number may be exponential, this does not lead to a further complexity blow up. Thus, Secure Planning is in NEXPTIME. On the other hand, even in plain planning domains, an exponential number of (exponentially long) candidate secure plans may exist, and the best we can do seems to be guessing a suitable one and verifying it.

**Effect of parallel actions** The results in Table 1 address the case where parallel actions in plans are allowed. However, excluding parallel actions and considering only sequential plans does not change the picture drastically. In all cases, the complexity stays the same except for secure planning under fixed plan length, where Secure Planning is $\Pi_2^P$-complete in general and $\mathrm{D}^P$-complete in proper planning domains (Theorem 5.7). Intuitively, this is explained by the fact that for a plan length fixed to a constant, the number of potential candidate plans is polynomially bounded in the input size of $\mathcal{P}$, and thus the guess of a proper secure candidate can be replaced by an exhaustive search, where it remains to check as a side issue the consistency of the domain (i.e., existence of some legal initial state), which is NP-complete in general (also for plain domains); see Theorem 5.7 below.

**Effect of nondeterministic actions** Our results also imply some conclusions on nondeterministic vs. deterministic planning domains. Interestingly, in proper planning domains, nondeterminism has no impact on the complexity for all problems considered, and we can conclude the same for Optimistic Planning as well as Secure Planning under arbitrary plan length. Furthermore, for proper planning problems even the combined restrictions of sequential plans and deterministic action outcomes do not decrease the complexity except for Secure Planning with fixed plan length, since the hardness results are obtained for plain planning problems, which guarantee these restrictions.

**Implications for implementation** The complexity results have important consequences for the implementation of $\mathcal{K}$ on top of existing computational logic systems, such as Blackbox [37], CCALC [47], smodels [33], DLV, satisfiability checkers, e.g. [53, 41, 2, 74], or Quantified Boolean Formula (QBF) solvers [4, 61, 18]. Optimistic Planning under arbitrary plan length is not polynomially reducible to systems with capability of solving problems within the Polynomial Hierarchy, e.g. Blackbox, satisfiability checkers, CCALC, smodels, or DLV, while it is feasible using QBF solvers. On the other hand, for fixed (and similarly, for polynomially bounded) plan length, Optimistic Planning can be polynomially expressed in all these systems. On the other hand, even in the case of fixed plan length and proper planning domains, Secure Planning is beyond the capability of systems having "only" NP expressiveness such as Blackbox, CCALC, smodels, or satisfiability checkers, while it can be encoded in DLV (which has $\Sigma_2^P$ expressiveness) and QBF solvers. Even in the more restrictive plain planning domains, where Secure Planning is $\mathrm{D}^P$-complete, the systems mentioned can not polynomially express Secure Planning in a single encoding. On the other hand, if we abandon properness, then also DLV is incapable of encoding Secure Planning (whose complexity increases to $\Sigma_3^P$-completeness). Nonetheless, Secure Planning is feasible in DLV using a two step approach as in [25], where optimistic plans are generated as secure candidate plans and then checked for security; this check is polynomially expressible in DLV.

Secure planning under arbitrary plan length is provably intractable, even in plain domains. Since NEXPTIME strictly contains PSPACE, there is no polynomial time transformation to QBF solvers or other popular computational logic systems with expressiveness limited to PSPACE, such as traditional STRIPS planning.

Here, further restrictions are needed to lower complexity to PSPACE, such as a polynomial bound on the plan length in the input query. We leave this for further investigation.

## 5 Derivation of Results

In this section, we show how the results discussed in the previous section are derived.

In the proofs of the lower bounds, the constructed planning problems $\mathcal{P} = \langle\langle \Pi, \langle D, R\rangle\rangle, q\rangle$ will always have empty background knowledge $\Pi$. Furthermore, the action and fluent declarations $F_D$ and $A_D$, respectively, will be as needed for the $R$-part, and are not explicitly mentioned. That is, we shall only explicitly address $R$ and $q$, while $\Pi = \emptyset$ and $D$ are implicitly understood.

The following lemma on checking initial states and legal state transitions is straightforward from well-known complexity results for logic programming (cf. [7]).

**Lemma 5.1** *Given a state $s_0$ (resp., a state transition $t = \langle s, A, s'\rangle$) and a propositional planning domain $PD = \langle \Pi, \langle D, R\rangle\rangle$, checking whether $s$ is a legal initial state (resp., $t$ is a legal state transition) is possible in polynomial time.*

**Proof**. [of Lemma 5.1] The unique answer set $M$ of the stratified normal logic program $\Pi$ can be computed in polynomial time (cf. [7]). Given $M$, the set of legal fluent and action instances $\mathcal{L}_{PD}$ is easily computable in polynomial time, as well as the reduction $PD^t$. Checking whether $s_0$ is a legal initial state for $PD^t$ amounts to checking whether $s_0$ is the least fix-point of a set of positive propositional rules, which is well-known polynomial. Overall, this means that checking whether $s_0$ is a legal initial state of $PD$ is polynomial. From $M$, $t$, and $PD^t$, it can be easily checked in polynomial time whether $A$ is executable w.r.t. $s$ and, furthermore, whether $s'$ is the minimal consistent set that satisfies all causation rules w.r.t. $s \cup A \cup M$ by computing the least fixpoint of a set of positive rules and verifying constraints on it. Thus, checking whether $t$ is a legal state transition is polynomial in the propositional case. □

**Corollary 5.2** *Given a sequence of state transitions $T = \langle t_1, \ldots, t_n\rangle$, where $t_i = \langle s_{i-1}, A_i, s_i\rangle$ for $i = 1, \ldots, n$, and a propositional planning domain $PD = \langle \Pi, \langle D, R\rangle\rangle$, checking whether $T$ is legal with respect to $PD$ is possible in polynomial time.*

## 5.1 Optimistic Planning

From the preparatory results, we thus obtain the following result on Optimistic Planning.

**Theorem 5.3** *Deciding whether for a given propositional planning problem $\mathcal{P} = \langle PD, q\rangle$ an optimistic plan exists is (a)* NP*-complete, if the plan length in $q$ is fixed, and (b)* PSPACE*-complete in general. The hardness parts hold even for plain $\mathcal{P}$.*

**Proof**. (a). The problem is in NP, since a trajectory $T = \langle t_1, \ldots, t_i\rangle$ where $t_j = \langle s_{j-1}, A_j, s_j\rangle$ for $j = 1, \ldots, i$, such that $s_i$ satisfies the goal $G$ in $q = G\ ?(i)$ can be guessed and, by Corollary 5.2, verified in polynomial time if $i$ is fixed.

NP-hardness for plain $\mathcal{P}$ is shown by a reduction from the satisfiability problem (SAT). Let $\phi = C_1 \wedge \cdots \wedge C_k$ be a CNF, i.e., a conjunction of clauses $C_i = L_{i,1} \vee \cdots \vee L_{i,m_i}$ where the $L_{i,j}$ are classical literals over propositional atoms $X = \{x_1, \ldots, x_n\}$. We declare these atoms and a further atom '0' as fluents in $D$, and put into the "`initially`" section $I_R$ of the planning domain $PD = \langle \emptyset, \langle D, R\rangle\rangle$ the following constraints:

| | |
|---|---|
| `total` $x_j$. | for all $x_j \in X$ |
| `forbidden` $-L_{i,1}, \ldots, -L_{i,m_i}$. | $1 \leq i \leq k$ |
| `caused` $0$. | |

Here, the first constraint effects the choice of a value for each fluent $x_j$. Clearly, $PD$ has a legal initial state iff $\phi$ is satisfiable. Thus, an optimistic plan $P$ exists for $\mathcal{P} = \langle PD, 0\ ?\ (0)\rangle$ iff $\phi$ is satisfiable. As $\mathcal{P}$ can easily be constructed from $\phi$, the result follows.

(b). A proof of membership in PSPACE follows from the discussion in Section 4.2 (note Lemma 5.1). We remark that the problem can be solved by a deterministic algorithm in polynomial workspace as follows. Similar as in [3], design a deterministic algorithm REACH$(s, s', \ell)$ which decides, given states $s$ and $s'$ and an integer $\ell$, whether a sequence $t_1, \ldots, t_\ell$ of legal transitions $t_i = \langle s_{i-1}, A_i, s_i\rangle$ exists, where $s = s_0$ and $s' = s_\ell$, by cycling trough all states $s''$ and recursively solving REACH$(s, s'', \lfloor \ell \rfloor)$ and REACH$(s'', s', \lfloor \ell \rfloor + 1)$. Then, the existence of an optimistic plan of length $\ell$ can be decided cyclic through all pairs of states $s, s'$ and testing whether $s$ is a legal initial state, $s''$ satisfies the goal in given in $q$, and REACH$(s, s', \ell)$ returns true. Since the recursion depth is $O(\log \ell)$, and each level of the recursion needs only polynomial space, Lemma 5.1 implies that this algorithm runs in polynomial space.

For the PSPACE-hardness part, we describe how propositional STRIPS planning as in [3] can be reduced to planning in $\mathcal{K}$, where the planning domain $PD$ is plain.

Recall that in propositional STRIPS, a state description $s$ is a consistent set of propositional literals, and an operator $op$ has a precondition $pc(op)$, an add-list $add(op)$, and a delete-list $del(op)$, which all are lists of propositional literals. The operator $op$ can be applied in $s$ if $pc(op) \subseteq s$ holds, and its execution yields the state $op(s) = (s \setminus del(op)) \cup add(op)$ (where $s'$ must be consistent). Otherwise, the application of $op$ on $s$ is undefined. A goal $\gamma$, which is a set of literals, can be reached from a state $s$, if there exists a sequence of operators $op_1, \ldots, op_\ell$, where $\ell \geq 0$, such that $s_i = op_i(s_{i-1})$, for $i = 1, \ldots, \ell$, where $s_0 = s$, and $\gamma \subseteq s_\ell$ holds. Any such sequence is called a ***STRIPS-plan*** (of length $\ell$) for $s, \gamma$. Given $s, \gamma$, a collection of STRIPS operators $op_1, \ldots, op_n$, and an integer $\ell \geq 0$, the problem of deciding whether some STRIPS-plan of length at most $\ell$ exists is PSPACE-complete [3]. As easily seen, this remains true if we ask for a plan of length exactly $\ell$ (just introduce a dummy operation with empty precondition and no effects).

Each STRIPS operator $op_i$ is easily modeled as action in language $\mathcal{K}$ using the following statements in the "`always`" section, i.e., the $C_R$ part of $R$:

`executable` $op_i$ `if` $pc(op_i)$.
`caused` $L$ `after` $op_i$. for each $L \in add(op_i) \setminus del(op_i)$
`caused` $L$ `after` $op_i$, $L$. for each $L \notin add(op_i) \cup del(op_i)$

The last rule is an inertia rule for the literals not affected by $op$.

The initial state $s$ of a STRIPS planning problem can be easily represented using the following constraints in the "`initially`" section, i.e., the $I_R$ part of $R$:

`caused` $L$. for all $L \in s$

Finally, $C_R$ contains the mandatory rules for unique action execution in a plain planning domain:

`nonexecutable` $op_i$ `if` $op_j$. $1 \leq i < j \leq n$
`caused false after not` $op_1$, `not` $op_2$, ..., `not` $op_n$.

It is easy to see that for the planning problem $\mathcal{P} = \langle PD, q\rangle$ where $PD = \langle \emptyset, AD\rangle$ and $q = \gamma\ ?\ (\ell)$, some optimistic plan exists iff a STRIPS-plan of length $\ell$ for $s, \gamma$ exists. Since $\mathcal{P}$ is constructible from the STRIPS instance in polynomial time, this proves the PSPACE-hardness part. □

## 5.2 Secure Planning

Secure Planning appears to be harder; already recognizing a secure plan is difficult.

**Theorem 5.4** *Given a propositional planning problem $\mathcal{P} = \langle PD, q\rangle$ and an optimistic plan $P$ for $\mathcal{P}$, deciding whether $P$ is secure is (a) $\Pi_2^P$-complete in general and (b) co-NP-complete, if $\mathcal{P}$ is proper.*[2] *Hardness in (a) and (b) holds even for fixed plan length in $q$ and sequential $P$, and if $\mathcal{P}$ in (b) is moreover plain.*

**Proof**. The plan $P = \langle A_1, \ldots, A_i\rangle$ for $\mathcal{P}$ is not secure, if a trajectory $T = \langle t_1, \ldots, t_\ell\rangle$, where $t_j = \langle s_{j-1}, A_j, s_j\rangle$, for $j = 1, \ldots, \ell$ exists, such that either (i) $\ell = i$ and $s_i$ does not satisfy the goal in $q$, or (ii) $\ell < i$ and for no state $s$, the tuple $\langle s_\ell, A_{\ell+1}, s\rangle$ is a legal transition. A trajectory $T$ of any length $\ell$ can, by Corollary 5.2, be guessed and verified in polynomial time. Condition (i) can be easily checked. Condition (ii) can be checked by a call to an NP oracle in polynomial time. It follows that checking security is in co-NP$^{\mathrm{NP}} = \Pi_2^P$ in general. If $\mathcal{P}$ is proper, condition (ii) can be checked in polynomial time, and thus the problem is in co-NP. This shows the membership parts.

$\Pi_2^P$-hardness in case (a) is shown by a reduction from deciding whether a QBF $\Phi = \forall X \exists Y \phi$ is true, where $X, Y$ are disjoint sets of variables and $\phi = C_1 \wedge \ldots \wedge C_k$ is a CNF over $X \cup Y$. It is well-known that this problem is $\Pi_2^P$-complete, cf. [54]. Without loss of generality, we assume that $\phi$ is satisfied if all atoms in $X \cup Y$ are set to true.

We declare the atoms in $X \cup Y$ and further atoms 0 and 1 as fluents in $D$. The "`initially`" section $I_R$ for $AD = \langle D, R\rangle$ has the following constraints:

`total` $x_j$. for all $x_j \in X$
`caused` 0.

The "`always`" section $C_R$ of $R$ contains the following rules. Suppose that $L_{i,1}, \ldots\ L_{i,n_i}$ are all literals over atoms from $X$ which occur in $C_i$, and similarly that $K_{i,1}, \ldots\ K_{i,m_i}$ are all literals over atoms from $Y$ that occur in $C_i$.

`total` $y_j$ `after` 0. for all $y_j \in X$
`forbidden` $-K_{i,1}, \ldots, -K_{i,m_i}$ `after` 0, $-L_{i,1}, \ldots, -L_{i,n_i}$. $1 \leq i \leq k$
`caused` 1 `after` 0.

These rules generate $2^{|X|}$ legal initial states $s_0^1, \ldots, s_0^{2^{|X|}}$ w.r.t. $\langle \emptyset, AD\rangle$, which correspond 1-1 to the truth assignments to the atoms in $X$. Each such $s_0^i$ contains precisely one of $x_j$ and $-x_j$, for all $x_j \in X$, and the atom 0. The totalization rule for $y_j$ effects that each legal state $s_1$ following the initial state contains exactly one of $y_j$ and $-y_j$. That is, $s_1$ must encode a truth assignment for $Y$. The `forbidden` statements check that the assignment to $X \cup Y$, given jointly by $s_0^i$ and $s_1$, satisfies all clauses of $\phi$. Furthermore, 1 must be contained in $s_1$ by the last rule.

Let us introduce an action $\alpha$, which is always executable. Then, the assumption on $\Phi$ implies that $P = \langle\langle s_0, A_1, s_1\rangle\rangle$, where $s_0 = X \cup \{0\}$, $A_1 = \{\alpha\}$, and $s_1 = X \cup Y \cup \{1\}$, is a trajectory w.r.t. $PD = \langle \emptyset, AD\rangle$, and thus $P = \langle A_1\rangle$ is an optimistic plan for the planning problem $\mathcal{P} = \langle PD, q\rangle$ where $q = 1\ ?\ (1)$. It is not hard to see that $P$ is secure iff $\Phi$ is true. Since $\langle PD, q\rangle$ is easily constructed from $\Phi$, this proves the hardness part of (a). The hardness part of (b) is established by a variant of the reduction; we disregard $Y$ (i.e., $Y = \emptyset$), and modify the rules as follows: `false` (after macro expansion) is replaced by

[2] We are grateful to Hudson Turner for pointing out that in a draft of [11], a co-NP-upper bound as reported there obtains only if deciding executability of an action (leading to a new legal state) is in P, and that the complexity in the general case may be one level higher up in PH. In fact, we were mainly interested in such domains, which are covered by our notion of proper domains.

1, and the rule with effect 1 is dropped. Note that the resulting planning domain is plain. Then, the plan $P = \langle A_1 \rangle$ is secure iff $\forall X \neg \phi$ is true, i.e., the CNF $\phi$ is unsatisfiable, which is co-NP-hard to check. □

For Secure Planning, we obtain the following result.

**Theorem 5.5** *Deciding whether a given propositional planning problem $\mathcal{P} = \langle PD, q \rangle$ has a secure plan is (a) $\Sigma_3^P$-complete, if the plan length in $q$ is fixed, (b) $\Sigma_2^P$-complete, if the plan length in $q$ is fixed and $\mathcal{P}$ is proper. Hardness in (b) holds even for deterministic and plain $PD$.*

**Proof**. a) and b). A trajectory $T = \langle \langle s_0, A_1, s_1 \rangle, \ldots, \langle s_{i-1}, A_i, s_i \rangle \rangle$ of fixed length $i$ that induces an optimistic plan $P = \langle A_1, \ldots, A_i \rangle$ can be guessed and verified in polynomial time (Corollary 5.2), and by Theorem 5.4, checking whether $P$ is secure is possible with a call to an oracle for $\Pi_2^P$ in case (a) and for co-NP in case (b). Hence, it follows that the problem is in $\Sigma_3^P$ in case (a) and in $\Sigma_2^P$ in case (b).

For the hardness part of (a), we transform deciding the validity of a QBF $\Phi = \exists Z \forall X \exists Y \phi$, where $X, Y, Z$ are disjoint sets of variables and $\phi = C_1 \ldots C_k$ is a CNF over $X \cup Y \cup Z$, which is $\Sigma_3^P$-complete [54], into this problem. The transformation extends the reduction in the proof of Theorem 5.4.

We introduce, for each atom $z_i \in Z$, an action $\mathtt{set}_{z_i}$ in $D$. The "`initially`" section, i.e., the $I_R$ part of $R$ contains the following constraints:

| | |
|---|---|
| `total` $x_j$. | for all $x_j \in X$ |
| `caused` 0. | |

$C_R$ contains the following rules. Suppose that $L_{i,1}, \ldots\ L_{i,n_i}$ are all literals over atoms from $X$ that occur in $C_i$, and similarly that $K_{i,1}, \ldots\ K_{i,m_i}$ are all literals over atoms from $Y \cup Z$ that occur in $C_i$.

| | |
|---|---|
| `caused` $z_i$ `after` 0, $\mathtt{set}_{z_i}$. | for all $z_i \in Z$ |
| `caused` $-z_i$ `after` 0, `not` $\mathtt{set}_{z_i}$. | for all $z_i \in Z$ |
| `caused` 1 `after` 0. | |
| `total` $y_j$ `after` 0. | for all $y_j \in Y$ |
| `forbidden` $-K_{i,1}, \ldots, K_{i,m_i}$ `after` 0, $-L_{i,1}, \ldots, -L_{i,n_i}$. | $1 \leq i \leq k$ |

Given these action descriptions, there are $2^{|X|}$ many legal initial states $s_0^1, \ldots, s_0^{2^{|X|}}$ for the emerging planning domain $PD = \langle \emptyset, AD \rangle$, which correspond 1-1 to the possible truth assignments to the variables in $X$ and contain 0. Executing in these states $s_0^i$ some actions $A$ means assigning a subset of $Z$ the value true. Every state $s_1^i$ reached from $s_0^i$ by a legal transition must, for each atom $\alpha \in Z \cup Y$, either contain $\alpha$ or $-\alpha$, where for the atoms in $Z$ this choice is determined by $A$. Furthermore, $s_1^i$ must contain the atom 1.

It is not hard to see that an optimistic plan of form $P = \langle A_1 \rangle$ (where $A_1 \subseteq \{\mathtt{set}_{z_i} \mid z_i \in Z\}$) for the goal 1 exists w.r.t. $PD = \langle \emptyset, AD \rangle$ iff there is an assignment to all variables in $X \cup Y \cup Z$ such that the formula $\phi$ is satisfied. Furthermore, $P$ is secure iff $A_1$ represents an assignment to the variables in $Z$ such that, regardless of which assignment to the variables in $X$ is chosen (which corresponds to the legal initial states $s_0^i$), there is some assignment to the variables in $Y$ (i.e., there is at least some state $s_1^i$ reachable from $s_0^i$, by doing $A_1$), such that all clauses of $\phi$ are satisfied; any such $s_1^i$ contains 1. In other words, $P$ is secure iff $\Phi$ is true.

Since $PD$ is constructible from $\Phi$ in polynomial time, it follows that deciding whether a secure plan exists for $\mathcal{P} = \langle PD, q \rangle$, where $q = 1$ ? (1), is $\Sigma_3^P$-hard. This proves part (a).

For the hardness part of (b), we modify the construction for part (a) by assuming that $Y = \emptyset$, and

- replace `false` in rule heads (after macro expansion) by 1;

- remove the rule for 1 and the `total`-rules for $y_j$).

The resulting planning domain $PD'$ is proper: since no causation rule in $C_R$ contains default negation, for each transition $t = \langle s, A, s_1\rangle$, the reduct ${PD'}^t$ coincides with ${PD'}^{\langle s,A,\emptyset\rangle}$, and thus existence of a a legal transition $\langle s, A, s_1\rangle$ can be determined in polynomial time. Furthermore, $\langle s, A, s_1\rangle$ is determined, and thus $PD'$ is also deterministic. We have again $2^{|X|}$ initial states $s_0^i$, which correspond to the truth assignments to $X$. An optimistic plan for the goal 1 of the form $P = \langle A_1\rangle$, where $A_1 \subseteq \{\mathtt{set}_{z_i} \mid z_i \in Z\}$, corresponds to an assignment to $Z \cup X$ such that $\phi$ evaluates to *false*. The plan $P$ is secure iff every assignment to $X$, extended by the assignment to $Z$ encoded by $A_1$, makes $\phi$ false.

It follows that a secure plan for $\mathcal{P} = \langle PD', q\rangle$, where $q = 1\ ?\ (1)$, exists iff the QBF $\exists Z \forall X \neg\phi$ is true. Evaluating a QBF of this form is $\Sigma_2^P$-hard (recall that $\phi$ is in CNF). Since $\mathcal{P}$ is constructible in polynomial time, this proves $\Sigma_2^P$-hardness for part (b). □

Next, we consider Secure Planning under arbitrary plan length.

As mentioned above, we can build a secure plan step by step only if we know all states that are reachable after the steps $A_1, \ldots, A_i$ so far when the next step $A_{i+1}$ is generated. Either we store these states explicitly, which needs exponential space in general, or we store the steps $A_1, \ldots, A_i$ (from which these states can be recovered) which also needs exponential space in the representation size of $\langle PD, q\rangle$. In any case, such a nondeterministic algorithm for generating a secure plan needs exponential time. The next result shows that NEXPTIME actually captures the complexity of deciding the existence of a secure plan.

**Theorem 5.6** *Deciding whether a given propositional planning problem $\mathcal{P} = \langle PD, q\rangle$ has a secure plan is* NEXPTIME-*complete. Hardness holds even for plain (and thus deterministic) $\mathcal{P}$.*

**Proof**. As for the membership part, the size of a string representing a secure plan $P = \langle A_1, \ldots, A_i\rangle$ of length $i$ for the query $q = \mathit{Goal}\ ?\ (i)$ is at most $O(i \cdot |PD|)$, which is single exponential in the sizes $|PD|$ and $\log i$ of the strings for $PD$ and $i$, respectively. Hence, this string has size single exponential in the size of $\mathcal{P}$. We can thus guess and verify a secure plan $P$ for $\mathcal{P}$ in (single) exponential time as follows:

1. Compute the set $\mathcal{S}_0$ of all legal initial states. If $\mathcal{S}_0 = \emptyset$, then $P$ is not secure (in fact, no secure plan exists).

2. Otherwise, for each $j = 1, \ldots, i$, compute for each $s \in \mathcal{S}_{j-1}$ the set $\mathcal{S}_j(s) = \{s' \mid \langle s, A_j, s'\rangle$ is a legal transition$\}$, and halt if some $\mathcal{S}_j(s)$ is empty; otherwise, set $\mathcal{S}_j = \bigcup_{s \in \mathcal{S}_{j-1}} \mathcal{S}_j(s)$.

3. Finally, check whether the goal is satisfied in every $s \in \mathcal{S}_i$, and accept iff this is true.

The computation of $\mathcal{S}_0$, as well as of each $\mathcal{S}_j(s)$, can be done in single exponential time, by considering all possible knowledge states $s'$ that might occur in a legal transition $\langle s, A_j, s'\rangle$. The number of different $\mathcal{S}_j(s)$ is exponentially bounded in the size of $\mathcal{P}$; thus, overall an exponential number of steps suffices to check whether the plan $P$ is secure.

The NEXPTIME-hardness part is shown by a generic Turing machine (TM) encoding. That is, given a nondeterministic TM $M$ which accepts a language $\mathcal{L}_{\mathcal{M}}$ in exponential time and an input word $w$, we show how to construct a plain planning problem $\mathcal{P} = \langle PD, q\rangle$ in polynomial time which has a secure plan iff $M$ accepts $w$. Roughly, the states in the set $\mathcal{S}_0$ of legal initial states encode the tape cells of $M$ and their initial contents; the actions in a secure plan represent the moves of the machine, which change the cell contents, and lead to acceptance of $w$. While the idea is clear, the technical realization bears some subtleties.

The reduction is as follows. Without loss of generality, $M$ halts on $w$ in less than $2^{n^k}$ many steps, where $n = |w|$ is the length of the input and $k \geq 0$ is some fixed integer (independent of $n$), and $M$ has a unique accepting state. We modify $M$ such that it loops in this state once it is reached. The cells $C_0, C_1 \ldots, C_N$, where $N = 2^{n^k} - 1$, of the work tape of $M$ (only those are relevant) are represented in different legal states of the planning domain. Initially, the cells $C_0, \ldots, C_{|w|-1}$ contain the symbols $w_0, w_1 \ldots, w_{|w|-1}$ of the input word $w$, and all other cells $C_{|w|}, \ldots, C_N$ are blank.

The computation of $M$ on $w$ is modeled by a secure plan $P = \langle A_1, \ldots, A_N \rangle$, in which each $A_j$ contains a single action $\alpha_{\tau_j}$ which models the transition of $M$ from the current configuration of the machine to the next one. A configuration of $M$, given by the contents of the work tape, the position of the read-write (rw) head, and the current state of the machine, is described by legal knowledge states $s_i$, $0 \leq i \leq N$, such that $s_i$ contains the symbol $\sigma$ currently stored in $C_i$, the current position $h$ of the rw-head, and the current state $q$ of $M$; all this information is encoded using fluents.

The information to which cell $C_i$ a legal knowledge state corresponds is given by literals $\pm i_1, \ldots, \pm i_{n^k}$, which represent the integer $i \in [0, N]$ in binary encoding, where $i_j$ (resp., $-i_j$) means that the $j$-th bit of $i$ is 1 (resp., 0). The position of the rw-head, $h \in [0, N]$, is represented similarly using further literals $\pm h_1, \ldots, \pm h_{n^k}$. Each symbol $\sigma$ in the tape alphabet $\Sigma$ of $M$ is represented by a fluent $p_\sigma$. Similarly, each state $q$ in the set $Q$ of states of $M$ is represented by a fluent $p_q$; in each legal knowledge state, exactly one $p_\sigma$ and one $p_q$ is contained. There are $2^{n^k}$ legal initial knowledge states, which uniquely describe the initial configuration of $M$, in which the rw-head of $M$ is placed over $C_0$, $M$ is in its initial state (say, $q_1$), and the work tape contains the input $w$.

The legal initial knowledge states $s$ are generated using constraints which "guess" a value for each bit of $i$, initialize the contents of $C_i$ with the right symbol $p_\sigma$, include $-h_j$ for all $j = 1, \ldots, n^k$ (i.e., set $h = 0$), and include $q_1$. More precisely, the "`initially`" section, i.e. $I_R$ of $R$ in $AD = \langle D, R \rangle$ is as follows:

| | |
|---|---|
| `total` $i_j$. | for all $j = 1, \ldots, n^k$ |
| `caused` $-h_j$. | for all $j = 1, \ldots, n^k$ % set $h = 0$ |
| `caused` $p_{w_0}$ `if` $-i_1, -i_2, \ldots, -i_{n^k}$. | % work tape position 0 |
| `caused` $p_{w_1}$ `if` $i_1, -i_2, \ldots, -i_{n^k}$. | % work tape position 1 |
| $\vdots$ | $\vdots$ |
| `caused` $p_{w_{\|w\|-1}}$ `if` "code of $\|w\| - 1$". | % work tape position $\|w\| - 1$ |
| `caused` $p_\sqcup$ `if not` $p_{\sigma_1}, \ldots,$ `not` $p_{\sigma_m}$. | % rest of tape is blank |
| `caused` $q_1$. | % initial state is $q_1$ |

Here, the tape alphabet $\Sigma$ is assumed to be $\Sigma = \{\sqcup, \sigma_1, \sigma_2, \ldots, \sigma_m\}$, where $\sqcup$ is the blank symbol.

The transition function of $M$ is given by tuples $\tau = \langle \sigma, q, \sigma', d, q' \rangle$, which reads as follows: if $M$ is in state $q$ and reads the symbol $\sigma$ at the current rw-head position $h$ (i.e., $C_h$ contains $\sigma$), then $M$ writes $\sigma'$ at the position $h$ (i.e., into $C_h$), moves the rw-head to position $h + d$, where $d = \pm 1$, and changes to state $q'$. (Without loss of generality, we omit here modeling that the rw-head might remain in the same position.)

Such a possible transition $\tau$ is modeled using rules which describe how to change a current knowledge state $s$, which corresponds to the tape cell $C_i$, to reflect $C_i$ in the new configuration of $M$. Informally, its constituents are manipulated as follows.

**work tape contents** For the case that $h = i$, i.e., the rw-head is at position $i$, a rule includes $p_\sigma$ into the state. Otherwise, i.e., the rw-head is not at $h$, an inertia rule includes $p_\sigma$, where $\sigma$ is the old contents of $C_i$, to the new knowledge state.

**rw-head position** The change of the rw-head position by $\pm 1$, is incorporated by replacing $h$ with $h \pm 1$. This is possible using a few rules, which simply realize an increment resp. decrement of the counter $h$. We assume at this point that $M$ is well-behaved, i.e., does not move left of $C_0$.

**state** A rule includes $p_{q'}$ for the resulting state $q'$ of $M$ into the new knowledge state.

To implement this, we introduce for each possible transition $\tau = \langle \sigma, q, \sigma', d, q' \rangle$ of $M$ an action $\alpha_\tau$, whose executability is stated in $C_R$ as follows:

`executable` $\alpha_\tau$ `if` $p_q,\ p_\sigma,\ h{=}i.$
`executable` $\alpha_\tau$ `if not` $h{=}i.$

Here $h{=}i$ is a *fluent* atom, which indicates whether the rw-head position $h$ is the index $i$ of the cell $C_i$ represented by the knowledge state.

Furthermore, several groups of rules are put in the "`always`" section, i.e. $C_R$ of $R$. The first group serves for determining the value of $h{=}i$, using auxiliary fluents $e_1, \ldots, e_{n^k}$:

`caused` $e_j$ `if` $h_j,\ i_j.$ for all $j = 1, \ldots, n^k$
`caused` $e_j$ `if` $-h_j,\ -i_j.$ for all $j = 1, \ldots, n^k$
`caused` $h{=}i$ `if` $e_1, \ldots, e_n^k.$

The execution of $\alpha_\tau$ effects a change in the state and the contents of $C_i$:

`caused` $p_{\sigma'}$ `after` $\alpha_\tau,\ h{=}i.$
`caused` $p_\sigma$ `after` $\alpha_\tau,\ p_\sigma,$ `not` $h{=}i.$ for all $\sigma \in \Sigma$
`caused` $p_{q'}$ `after` $\alpha_\tau.$

Depending on the value of $d$, different rules are added for realizing the move of the rw-head. Recall that, given the binary representation $x011\cdots1$ of an integer $z$, the binary representation of $z+1$ is $x100\cdots0$. The rules for $d = 1$ are as follows.

`caused` $h_1$ `after` $\alpha_\tau,\ -h_1.$
`caused` $h_2$ `after` $\alpha_\tau,\ -h_2,\ h_1.$
`caused` $-h_1$ `after` $\alpha_\tau,\ -h_2,\ h_1.$
$\vdots$
`caused` $h_{n^k}$ `after` $\alpha_\tau,\ -h_{n^k},\ h_{n^k-1}, \ldots, h_1.$
`caused` $-h_{n^k-1}$ `after` $\alpha_\tau,\ -h_{n^k},\ h_{n^k-1}, \ldots, h_1.$
$\ldots$
`caused` $-h_1$ `after` $\alpha_\tau,\ -h_{n^k},\ h_{n^k-1}, \ldots, h_1.$
`caused` $h_\ell$ `after` $\alpha_\tau,\ h_\ell,\ -h_j.$ where $1 \leq j < \ell \leq n^k$
`caused` $-h_\ell$ `after` $\alpha_\tau,\ -h_\ell,\ -h_j.$ where $1 \leq j < \ell \leq n^k$

The last two rules serve for carrying the leading bits of $i$, which are not affected by the increment, over to the new knowledge state. (This could also be realized in a simpler way using `inertial` statements; however, recall that such rules are not allowed in plain domains.)

The rules for $d = -1$ are similar, with the roles of 0 and 1 interchanged:

caused $-h_1$ after $\alpha_\tau,\ h_1$.
caused $-h_2$ after $\alpha_\tau,\ h_2,\ -h_1$.
caused $h_1$ after $\alpha_\tau,\ h_2,\ -h_1$.
$\vdots$
caused $-h_{n^k}$ after $\alpha_\tau,\ h_{n^k},\ -h_{n^k-1},\ldots,-h_1$.
caused $h_{n^k-1}$ after $\alpha_\tau,\ h_{n^k},\ -h_{n^k-1},\ldots,-h_1$.
$\ldots$
caused $h_1$ after $\alpha_\tau,\ h_{n^k},\ -h_{n^k-1},\ldots,-h_1$.
caused $h_\ell$ after $\alpha_\tau,\ h_\ell,\ h_j$. where $1 \leq j < \ell \leq n^k$
caused $-h_\ell$ after $\alpha_\tau,\ -h_\ell,\ h_j$. where $1 \leq j < \ell \leq n^k$

Further rules are added to $C_R$ for carrying the cell index $i$ over to the next knowledge state:

caused $i_j$ after $i_j$. for all $j = 1,\ldots,n^k$
caused $-i_j$ after $-i_j$. for all $j = 1,\ldots,n^k$

Finally, the mandatory rules of a plain planning domain enforcing the execution of one and only one action in each transition are added to $C_R$.

As easily checked, all rules that we have introduced satisfy the syntactic restrictions for plain planning domains.

Suppose now that $q_m \in Q$ is the unique accepting state of $M$. Then, a secure plan $P = \langle A_1,\ldots,A_\ell\rangle$ of length $\ell$ reaching the goal $q_m$ corresponds to the fact that $M$ will, starting from the initial configuration, be in an accepting configuration after executing the transitions $\tau_1,\ldots,\tau_\ell$, where $A_j = \{\alpha_{\tau_j}\}$, for $j = 1,\ldots,\ell$. By our assumption on $M$, we know that $M$ can reach some accepting configuration within $N$ steps iff it can reach an accepting configuration in exactly $N$ steps. Thus, we have that $M$ accepts the input $w$ iff there exists some secure plan of length $N$ for the goal $q_m$ in the planning domain $PD = \langle\emptyset, AD\rangle$ where $AD$ is from above. In other words, $M$ accepts $w$ within $N$ steps iff the proper propositional planning problem $\mathcal{P} = \langle PD,\ q_m\ ?\ (N)\rangle$ has a secure plan.

As easily seen, $\mathcal{P}$ can be constructed in polynomial time from $M$ and $w$. This proves NEXPTIME-hardness of deciding the existence of a secure plan, even under the restriction to plain planning problems. $\Box$

Secure planning has lower complexity if the plan length is fixed and concurrent actions are not allowed.

**Theorem 5.7** *Deciding whether a given propositional planning problem $\mathcal{P} = \langle PD, q\rangle$ has a secure sequential plan is (a) $\Pi_2^P$-complete, if $q$ is fixed, and (b) $\mathrm{D}^P$-complete, if $q$ is fixed and $\mathcal{P}$ is proper. The hardness part of (b) holds even for plain $\mathcal{P}$.*

**Proof**. If the plan length $i$ in the query $q = Goal\ ?\ (i)$ is fixed, the number of candidate sequential secure plans, given by $(a+1)^i$, where $a$ is the number of actions in $PD$, is bounded by a polynomial.

A candidate $P = \langle A_1,\ldots,A_n\rangle$ is not a secure plan, if (i) no initial state $s_0$ exists, or (ii) like in the proof of Theorem 5.4, a trajectory $T = \langle t_1,\ldots,t_\ell\rangle$, where $t_j = \langle s_{j-1}, A_j, s_j\rangle$, for $j = 1,\ldots,\ell$ exists, such that either (ii.1) $\ell = i$ and $s_i$ does not satisfy the goal in $q$, or (ii.2) $\ell < i$ and for no state $s$, the tuple $\langle s_\ell,\ A_{\ell+1}, s\rangle$ is a legal transition. The test for (i) is in co-NP, while the test for (ii) is in $\Sigma_2^P$ in general and in NP if $\mathcal{P}$ is proper (cf. proof of Theorem 5.4). Note that (i) is identical for all candidates.

Thus, the existence of a sequential secure plan can be decided by the conjunction of a problem in NP and a disjunction of polynomially many instances of a problem in $\Pi_2^P$ in case (a) and in co-NP in case (b);

since NP $\subseteq \Pi_2^P$ and both $\Pi_2^P$ and co-NP are closed under polynomial disjunctions and conjunctions of instances (i.e., a logical disjunction resp. conjunction of instances can be polynomially transformed into an equivalent single instance), it follows that the problem is in $\Pi_2^P$ in case (a) and in $\mathrm{D}^P$ in case (b).

$\Pi_2^P$-hardness for case (a) follows from the reduction in the proof of Theorem 5.4. There, a secure, sequential plan exists for the query $1 ~?~ (1)$ iff the plan $P = \langle\{\alpha\}\rangle$ is the secure.

$\mathrm{D}^P$-hardness for case (b) is shown by a reduction from deciding, given CNFs $\phi = \bigwedge_{i=1}^{n} L_{i,1} \vee L_{i,2} \vee L_{i,3}$ and $\psi = \bigwedge_{j=1}^{m} K_{j,1} \vee K_{j,2} \vee K_{j,3}$ over disjoint sets of atoms $X$ and $Y$, respectively, whether $\phi$ is satisfiable and $\psi$ is unsatisfiable.

The "`initially`" section, i.e., $I_R$ of $R$ contains the following constraints:

| | |
|---|---|
| `total` $x_j$. | for all $x_j \in X$ |
| `caused` $L_{i,1}$ `if` $-L_{i,2},\ -L_{i,3}$. | for all $i = 1, \ldots, n$ |
| `total` $y_j$. | for all $y_j \in Y$ |
| `caused` $f$ `if` $-K_{i,1},\ -K_{i,2},\ -K_{i,3}$. | for all $i = 1, \ldots, m$ |

Obviously, these rules satisfy the conditions for a plain planning domain. Then, for the query $q = f ~?~ (0)$, the only candidate for a sequential secure plan is the empty plan $P = \langle\rangle$. As easily seen, $P$ is a secure plan for $q$ iff $\phi$ is satisfiable (which is equivalent to the existence of some legal initial state) and $\psi$ is unsatisfiable (which means that $f$ is true in each initial state). This proves the hardness part of (b). □

We conclude this section with remarking that the constructions in the proofs of the hardness parts of Theorem 5.4, items (a) and (b) of Theorem 5.5, and item (a) of Theorem 5.7 involve planning problems that have length fixed to 1. For plan length fixed to 0, these problems have lower complexity (co-NP-completeness for the problems in Theorem 5.4 and $\mathrm{D}^P$-completeness for the other problems).

# 6 Related Work

There is a huge body of literature on planning (see [72, 73] for surveys). We will only focus on directly related research:

- Action languages and answer set planning
- Planning under incomplete knowledge
- Planning Complexity

## 6.1 Action Languages and Answer Set Planning

The language $\mathcal{K}$ proposed in this paper builds on earlier work on action languages [24]. The language $\mathcal{A}$, proposed in [23] provides a rudimentary set of causal statements, which roughly corresponds to $\mathcal{K}$ with complete states in which all rules $r$ are of the form (2) of section 2.1 with $\mathsf{post}(r) = \emptyset$, all actions are executable by default in any state, and all fluents are inertial. The language $\mathcal{B}$ described in [24] is very similar to $\mathcal{A}$, the difference is that the restriction on rules is relaxed and rules $r$ of the form (2) with $\mathsf{pre}(r) = \emptyset$ are allowed additionally, enabling the formulation of ramifications.

The language $\mathcal{C}$, proposed in [27] and based on the theory of causal explanation in [48, 42], is the action language which is closest to $\mathcal{K}$. In $\mathcal{C}$ not all fluents are automatically inertial – just as in $\mathcal{K}$ it must be explicitly declared if a fluent has the property of being inertial. As in $\mathcal{K}$, this is achieved by a macro inertial F. which is defined in $\mathcal{C}$ as caused F if F after F. whereas in $\mathcal{K}$ it is defined as caused F if not $-$ F after F. Furthermore, $\mathcal{C}$ has like $\mathcal{K}$ a macro default F. for declaring that a property holds by default. In $\mathcal{C}$, it stands for caused F if F, while in $\mathcal{K}$, it is defined as caused F if not $-$ F. The difference in macro expansion is due to the different semantic definition of transitions and also due to the lack of default negation in $\mathcal{C}$. In particular, default F. means in $\mathcal{C}$ that F is true without the need of further causal support. Finally, $\mathcal{C}$ also provides a way to specify nondeterministic action effects.

None of the abovementioned languages explicitly supports initial state constraints, nor does any support explicit executability conditions. Most importantly, their underlying semantics is not based on knowledge states, so fluents may not be undefined in any state. As a consequence, totality of fluents cannot be expressed in any of the languages $\mathcal{A}, \mathcal{B}$, and $\mathcal{C}$, as each fluent is implicitly total, and default negation is not supported.

In [65, 8] two approaches can be found, in which planning problems are formulated directly using answer set programming, without an intermediate representation in an action language. These approaches have an obvious representational deficiency, as recurring patterns and concepts are not summarized in a more abstract action language. The problems studied in these papers do not contain ramifications, and all fluents are assumed to be inertial; explicit executability conditions are considered, though. Furthermore, none of these approaches comprises nondeterministic action effects or incomplete initial states. Default negation is only used for the implementation of the planning framework and is not allowed for the specification of the transition system.

## 6.2 Planning Under Incomplete Knowledge

Planning under incomplete knowledge has been widely investigated in the AI literature. Most works extend algorithms/systems for classical planning, rather than using deduction techniques for solving planning tasks as proposed in this paper. The systems Buridan [39], UDTPOP [55], Conformant Graphplan [64], CNLP [56] and CASSANDRA [58] fall in this class. In particular, Buridan, UDTPOP, and Conformant Graphplan can solve secure planning (also called conformant planning) like $\text{DLV}^{\mathcal{K}}$. On the other hand, the systems CNLP and CASSANDRA deal with conditional planning (where the sequence of actions to be executed depends on dynamic conditions).

More recent works propose the use of automated reasoning techniques for planning under incomplete knowledge. In [60] a technique for encoding conditional planning problems in terms of 2-QBF formulas is proposed. The work in [21] proposes a technique based on regression for solving secure planning problems in the framework of the situation calculus, and presents a Prolog implementation of such a technique. In [49], sufficient syntactic conditions ensuring security of every (optimistic) plan are singled out. While sharing their logic-based nature, our work presented in this paper differs considerably from such proposals, since it is based on a different formalism.

Work similar to ours has been independently reported in [25]. In that paper, the author presents a SAT-based procedure for computing secure plans over planning domains specified in the action language $\mathcal{C}$ [27, 43, 45]. The main differences between our paper and [25] are (i) the different action languages used for specifying planning domains: $\mathcal{C}$ vs $\mathcal{K}$; the former is closer to classical logic, while the latter is more "logic programming oriented" by the use default negation; (ii) the different computational engines underlying the two systems (a SAT Checker vs a DLP system), which imply completely different translation techniques for the implementation.

### 6.3 Planning Complexity

Our results on the complexity of planning in $\mathcal{K}$ are related to several results in the planning literature. First and foremost, planning in STRIPS can be easily emulated in $\mathcal{K}$ planning domains, and thus results for STRIPS planning carry over to respective planning problems in $\mathcal{K}$, in particular Optimistic Planning, which by the results in [3, 14] is PSPACE-complete.

As for finding secure plans (alias conformant or valid plans), there have been interesting results in the recent literature. Turner [69] has analyzed in a recent paper the effect of various assumptions on different planning problems, including conformant planning and conditional planning under domain representation based on classical propositional logic. In particular, Turner reports that deciding the existence of a classical (i.e., optimistic) plan of polynomial length is $\mathrm{NP}$-complete, and $\mathrm{NP}$-hard already for length 1 where actions are always executable. Furthermore, he reports that deciding the existence of a conformant (i.e., secure) plan of polynomial length is $\Sigma_3^P$-complete, and $\Sigma_3^P$-hard already for length 1. Furthermore, the problem is reported $\Sigma_2^P$-complete if, in our terminology, the planning domain is proper, and $\Sigma_2^P$-hard for length 1 in deterministic planning domains. Turner's results match our complexity results, announced in [11]; this is intuitively sound, since answer set semantics and classical logic, which underlies ours and his framework, respectively, have the same computational complexity.

Enrico Giunchiglia [25] considered conformant planning in the action language $\mathcal{C}$, where concurrent actions, constraints on the action effects, and nondeterminism on both the initial state and effects of the actions are allowed – all these features are provided in our language $\mathcal{K}$ as well. Furthermore, Giunchiglia presented the planning system $\mathcal{C}$-plan, which is based on SAT solvers for computing, in our terminology, optimistic and secure plans following a two step approach. For this purpose, transformations of finding optimistic plans and security checking into SAT instances and QBFs are provided. The same approach is studied in [19] for an extension of STRIPS in which part of the action effects may be nondeterministic. While not explicitly analyzed, the structures of the QBFs emerging in [25, 19] reflect our complexity results for Optimistic Planning and Security Checking.

Rintanen [60] considered planning in a STRIPS-style framework. He showed that, in our terminology, deciding the existence of a polynomial-length sequential optimistic plan for every totalization of the initial state, given that actions are deterministic, is $\Pi_2^P$-complete. Furthermore, Rintanen showed how to extract a *single* such plan $P$ which works for all these totalizations, from an assignment to the variables $X$ witnessing the truth of a QBF $\exists X \forall Y \exists Z\, \phi$ that is constructed in polynomial time from the planning instance. Thus, the associated problem of deciding whether such a plan $P$ exists is in $\Sigma_3^P$. Note that intuitively, checking suitability of a given optimistic plan is in this problem more difficult than Security Checking, since only the operability of some trajectory vs all trajectories must be checked for each initial state. However, the problems have the same complexity ($\Pi_2^P$-hardness for Rintanen's problem is obtained by slightly adapting the proof of Theorem 5.4), and are thus polynomially intertranslatable. Following Rintanen's and Giunchiglia's approach, finding secure plans for planning problems in $\mathcal{K}$ can be mapped to solving QBFs. However, since our framework is based on answer set semantics, the respective QBFs will be more involved due to intrinsic minimality conditions of the answer set semantics.

Baral et al. [1] studied the complexity of planning under incomplete information about initial states in the language $\mathcal{A}$ [23], which is similar to the framework in [60] and gives rise to proper, deterministic planning domains. They show that deciding the existence of an, in our terminology, polynomial-length secure sequential plan is $\Sigma_2^P$-complete. Notice that we have considered this problem for plans of fixed length, for which this problem is $\mathrm{D}^P$-complete and thus simpler.

From our results on the complexity of planning in the language $\mathcal{K}$, similar complexity results may be derived for other declarative planning languages, such as STRIPS-like formalisms as in [60] and the language $\mathcal{A}$ [23], or the fragment of $\mathcal{C}$ restricted to causation of literals (cf. [25]), by adaptations of our complexity proofs. The intuitive reason is that in all these formalisms, state transitions are similar in spirit and have similar complexity characteristics. In particular, our results on Secure Planning should be easily transferred to these formalisms by adapting our proofs for the appropriate problem setting.

# 7 Conclusion

In this paper, we have presented an approach to knowledge-state planning, based on nonmonotonic logic programming. We have introduced the language $\mathcal{K}$, defined its syntax and semantics, and then shown how this language can be used to represent various planning problems from the planning literature, in various settings comprising incomplete initial states, nondeterministic actions effects, and parallel executions of actions. In particular, we have shown how knowledge-states, rather then world states, can be used in representing planning problems. We then have thoroughly analyzed the computational complexity of propositional planning problems in $\mathcal{K}$, where we have considered optimistic planning and secure (i.e., conformant) planning. As we have seen, under various restrictions these problems range in complexity from the first level of the Polynomial Hierarchy to NEXPTIME. In particular, secure planning under fixed vs variable plan length turned out to be $\Sigma_3^P$-complete and NEXPTIME-complete, respectively. Finally, we have compared our work to a number of related planning approaches and complexity results from the literature.

As we believe, the language $\mathcal{K}$, and in particular the nonmonotonic negation operator available in it, allows for a more convenient and natural representation of certain pieces of knowledge that are part of a planning problem than similar languages. In particular, this applies to Giunchiglia and Lifschitz's important language $\mathcal{C}$, which was the starting point for developing our $\mathcal{K}$ language. We have demonstrated that natural knowledge-state encodings of particular planning problems, e.g. some versions of the "bomb in the toilet" problem, exist, for which the problem of finding optimistic plans coincides with the problem of finding secure plans, while for encodings in the literature, which are based on the world state paradigm, this equivalence does not hold — all of the world-state-based encodings require secure planning, which is conceptually and computationally harder. We point out that the "bomb in the toilet" problems per se are computationally easy, so it seems that encodings based on world states artificially bloat these problems because of their lack of allowing a natural statement about fluents being unknown in some state.

Indeed, we have verified experimentally, using the $\text{DLV}^{\mathcal{K}}$ system, that the knowledge-state encodings of the "bomb in the toilet" problems reported in this paper run considerably faster than their world-state-based counterparts. The $\text{DLV}^{\mathcal{K}}$ system, which is described in detail in a companion paper [12], implements the language $\mathcal{K}$ on top of the DLV logic programming system [13, 16]. It supports both optimistic and secure planning (currently, the latter is supported for restricted classes of planning problems). Extensive experimental evaluation has shown that the $\text{DLV}^{\mathcal{K}}$ system, even if it was built merely as a front end to another system and not optimized for performance, had reasonable performance compared to other similar systems, and even outperformed various specialized systems for conformant planning under the use of knowledge-state problem encodings. This shows that nonmonotonic logic programming has potential for declarative planning, and that, in our opinion, further exploration of the knowledge-state encoding approach is worthwhile to pursue from a computational perspective.

While we have presented the language $\mathcal{K}$ and discussed its basic features and advantages, several issues are currently investigated or scheduled for future work. As for the implementation, we have already mentioned the $\text{DLV}^{\mathcal{K}}$ system, which will be improved in a steady effort. An intriguing issue in that is the design of efficient algorithms and methods for secure planning, since this problem is rather complex even for short plans (it resides at the third level of the Polynomial Hierarchy). Furthermore, we are currently exploring a possible enhancement of the planning formalism to computing optimal plans, i.e., plans whose execution cost, measured in accumulated costs of primitive action execution, is smallest over all plans. An implementation of optimal planning may take advantage of DLV's optimization features which are available through weak constraints. Finally, extensions of the language by further constructs such as sensing operators are part of future work.

**Acknowledgments** This work has greatly benefited from interesting discussions with and comments of Michael Gelfond, Vladimir Lifschitz, Riccardo Rosati, and Hudson Turner. Furthermore, we are grateful to Claudio Castellini, Alessandro Cimatti, Esra Erdem, Enrico Giunchiglia, David E. Smith, and Dan Weld for kindly supplying explanations, support, and comments on the systems that we used for comparison.

# References

[1] C. Baral, V. Kreinovich, and R. Trejo. Computational complexity of planning and approximate planning in the presence of incompleteness. *Artificial Intelligence*, 122(1-2):241–267, 2000.

[2] R. Bayardo and R. Schrag. Using CSP look-back techniques to solve real-world SAT instances. In *Proceedings of the 15th National Conference on Artificial Intelligence (AAAI-97)*, pages 203–208, 1997.

[3] T. Bylander. The Computational Complexity of Propositional STRIPS Planning. *Artificial Intelligence*, 69:165–204, 1994.

[4] M. Cadoli, A. Giovanardi, and M. Schaerf. An Algorithm to Evaluate Quantified Boolean Formulae. In *Proceedings AAAI/IAAI-98*, pages 262–267, 1998.

[5] A. Cimatti and M. Roveri. Conformant Planning via Model Checking. In *Proceedings of the Fifth European Conference on Planning (ECP'99)*, pages 21–34, September 1999.

[6] A. Cimatti and M. Roveri. Conformant Planning via Symbolic Model Checking. *Journal of Artificial Intelligence Research*, 13:305–338, 2000.

[7] E. Dantsin, T. Eiter, G. Gottlob, and A. Voronkov. Complexity and expressive power of logic programming. In *Proceedings of the Twelfth Annual IEEE conference on Computational Complexity, June 24–27, 1997, Ulm, Germany, (CCC'97)*, pages 82–101. Computer Society Press, June 1997. Extended paper in *ACM Computing Surveys*, 33(3), September 2001.

[8] Y. Dimopoulos, B. Nebel, and J. Koehler. Encoding Planning Problems in Nonmonotonic Logic Programs. In *Proceedings of the European Conference on Planning 1997 (ECP-97)*, pages 169–181. Springer Verlag, 1997.

[9] J. Dix. Semantics of Logic Programs: Their Intuitions and Formal Properties. An Overview. In *Logic, Action and Information. Proc. of the Konstanz Colloquium in Logic and Information (LogIn'92)*, pages 241–329. DeGruyter, 1995.

[10] P. M. Dung. On the relations between stable and well-founded semantics of logic programs. *Theoretical Computer Science*, 105(1):7–25, 1992.

[11] T. Eiter, W. Faber, N. Leone, G. Pfeifer, and A. Polleres. Planning under incomplete knowledge. In J. Lloyd, V. Dahl, U. Furbach, M. Kerber, K.-K. Lau, C. Palamidessi, L. M. Pereira, Y. Sagiv, and P. J. Stuckey, editors, *Computational Logic - CL 2000, First International Conference, Proceedings*, number 1861 in Lecture Notes in AI (LNAI), pages 807–821, London, UK, July 2000. Springer Verlag.

[12] T. Eiter, W. Faber, N. Leone, G. Pfeifer, and A. Polleres. A Logic Programming Approach to Knowledge-State Planning, II: the DLV$^{\mathcal{K}}$ System. Manuscript, November 2001.

[13] T. Eiter, N. Leone, C. Mateis, G. Pfeifer, and F. Scarcello. The KR System `dlv`: Progress Report, Comparisons and Benchmarks. In A. G. Cohn, L. Schubert, and S. C. Shapiro, editors, *Proceedings Sixth International Conference on Principles of Knowledge Representation and Reasoning (KR'98)*, pages 406–417. Morgan Kaufmann Publishers, 1998.

[14] K. Erol, D. S. Nau, and V. Subrahmanian. Complexity, decidability and undecidability results for domain-independent planning. *Artificial Intelligence*, 76(1-2):75–88, July 2000.

[15] K. Eshghi. Abductive Planning with Event Calculus. In *Proc. 5th International Conference and Symposium on Logic Programming*, pages 562–579. MIT Press, 1988.

[16] W. Faber, N. Leone, and G. Pfeifer. Pushing Goal Derivation in DLP Computations. In M. Gelfond, N. Leone, and G. Pfeifer, editors, *Proceedings of the 5th International Conference on Logic Programming and Nonmonotonic Reasoning (LPNMR'99)*, number 1730 in Lecture Notes in AI (LNAI), pages 177–191, El Paso, Texas, USA, December 1999. Springer Verlag.

[17] F. Fages. Consistency of clark's completion and existence of stable models. *Journal of Methods of Logic in Computer Science*, 1(1):51–60, 1994.

[18] R. Feldmann, B. Monien, and S. Schamberger. A Distributed Algorithm to Evaluate Quantified Boolean Formulae. In *Proceedings National Conference on AI (AAAI'00)*, pages 285–290, Austin, Texas, July 30-August 3 2000. AAAI Press.

[19] P. Ferraris and E. Giunchiglia. Planning as Satisfiability in Nondeterministic Domains. In *Proceedings of the Seventeenth National Conference on Artificial Intelligence (AAAI'00), July 30 – August 3, 2000, Austin, Texas USA*, pages 748–753. AAAI Press / The MIT Press, 2000.

[20] R. E. Fikes and N. J. Nilsson. Strips: A new approach to the application of theorem proving to problem solving. *Artificial Intelligence*, 2(3-4):189–208, 1971.

[21] A. Finzi, F. Pirri, and R. Reiter. Open world planning in the situation calculus. In *Proceedings of the Seventeenth National Conference on Artificial Intelligence (AAAI'00), July 30 – August 3, 2000, Austin, Texas USA*, pages 754–760. AAAI Press / The MIT Press, 2000.

[22] M. Gelfond and V. Lifschitz. Classical Negation in Logic Programs and Disjunctive Databases. *New Generation Computing*, 9:365–385, 1991.

[23] M. Gelfond and V. Lifschitz. Representing Action and Change by Logic Programs. *Journal of Logic Programming*, 17:301–321, 1993.

[24] M. Gelfond and V. Lifschitz. Action languages. *Electronic Transactions on Artificial Intelligence*, 2(3-4):193–210, 1998.

[25] E. Giunchiglia. Planning as Satisfiability with Expressive Action Languages: Concurrency, Constraints and Nondeterminism. In A. G. Cohn, F. Giunchiglia, and B. Selman, editors, *Proceedings of the Seventh International Conference on Principles of Knowledge Representation and Reasoning (KR 2000), April 12-15, Breckenridge, Colorado, USA*, pages 657–666. Morgan Kaufmann, 2000.

[26] E. Giunchiglia, G. N. Kartha, and V. Lifschitz. Representing action: Indeterminacy and ramifications. *Artificial Intelligence*, 95:409–443, 1997.

[27] E. Giunchiglia and V. Lifschitz. An Action Language Based on Causal Explanation: Preliminary Report. In *Proceedings of the Fifteenth National Conference on Artificial Intelligence (AAAI '98)*, pages 623–630, 1998.

[28] E. Giunchiglia and V. Lifschitz. Action languages, temporal action logics and the situation calculus. In *Working Notes of the IJCAI'99 Workshop on Nonmonotonic Reasoning, Action, and Change*, 1999.

[29] R. Goldman and M. Boddy. Expressive planning and explicit knowledge. In *Proceedings AIPS-96*, pages 110–117. AAAI Press, 1996.

[30] G. Gottlob, N. Leone, and H. Veith. Succinctness as a Source of Expression Complexity. *Annals of Pure and Applied Logic*, 97(1–3):231–260, 1999.

[31] C. C. Green. Application of theorem proving to problem solving. In *Proceedings IJCAI '69*, pages 219–240, 1969.

[32] S. Hanks and D. McDermott. Nonmonotonic logic and temporal projection. *Artificial Intelligence*, 33(3):379–412, 1987.

[33] Ilkka Niemelä. Logic programming with stable model semantics as constraint programming paradigm. *Annals of Mathematics and Artificial Intelligence*, 25(3–4):241–273, 1999.

[34] L. Iocchi, D. Nardi, and R. Rosati. Planning with sensing, concurrency, and exogenous events: Logical framework and implementation. In A. G. Cohn, F. Giunchiglia, and B. Selman, editors, *Proceedings of the Seventh International Conference on Principles of Knowledge Representation and Reasoning (KR 2000), April 12-15, Breckenridge, Colorado, USA*, pages 678–689. Morgan Kaufmann Publishers, Inc., 2000.

[35] G. N. Kartha and V. Lifschitz. Actions with indirect effects (preliminary report). In *Proceedings of the Fourth International Conference on Principles of Knowledge Representation and Reasoning (KR 94)*, pages 341–350, 1994.

[36] H. Kautz and B. Selman. Planning as Satisfiability. In *Proceedings of the 10th European Conference on Artificial Intelligence (ECAI '92)*, pages 359–363, 1992.

[37] H. Kautz and B. Selman. Unifying sat-based and graph-based planning. In *The International Joint Conferences on Artificial Intelligence (IJCAI) 1999*, pages 318–325, Stockholm, Sweden, Aug. 1999.

[38] R. Kowalski and M. Sergot. A logic-based calculus of events. *New Generation Computing*, 4:67–95, 1986.

[39] N. Kushmerick, S. Hanks, and D. S. Weld. An algorithm for probabilistic planning. *Artificial Intelligence*, 76(1–2):239–286, 1995.

[40] H. J. Levesque, R. Reiter, Y. Lespérance, F. Lin, and R. B. Scherl. GOLOG: A logic programming language for dynamic domains. *Journal of Logic Programming*, 31(1–3):59–83, 1997.

[41] C. Li and Anbulagan. Heuristics based on unit propagation for satisfiability problems. In *Proceedings of the Fourteen International Joint Conference on Artificial Intelligence (IJCAI) 1997*, pages 366–371, Nagoya, Japan, August 1997.

[42] V. Lifschitz. On the logic of causal explanation. *Artificial Intelligence*, 96:451–465, 1997.

[43] V. Lifschitz. Action Languages, Answer Sets and Planning. In K. Apt, V. W. Marek, M. Truszczyński, and D. S. Warren, editors, *The Logic Programming Paradigm – A 25-Year Perspective*, pages 357–373. Springer Verlag, 1999.

[44] V. Lifschitz. Answer set planning. In D. D. Schreye, editor, *Proceedings of the 16th International Conference on Logic Programming (ICLP'99)*, pages 23–37, Las Cruces, New Mexico, USA, Nov. 1999. The MIT Press.

[45] V. Lifschitz and H. Turner. Representing transition systems by logic programs. In M. Gelfond, N. Leone, and G. Pfeifer, editors, *Proceedings of the 5th International Conference on Logic Programming and Nonmonotonic Reasoning (LPNMR'99)*, number 1730 in Lecture Notes in AI (LNAI), pages 92–106, El Paso, Texas, USA, December 1999. Springer Verlag.

[46] W. Marek and M. Truszczyński. Autoepistemic Logic. *Journal of the ACM*, 38(3):588–619, 1991.

[47] N. McCain. The clausal calculator homepage, 1999. `<URL:http://www.cs.utexas.edu/users/tag/cc/>`.

[48] N. McCain and H. Turner. Causal theories of actions and change. In *Proceedings of the 15th National Conference on Artificial Intelligence (AAAI-97)*, pages 460–465, 1997.

[49] N. McCain and H. Turner. Satisfiability planning with causal theories. In A. G. Cohn, L. Schubert, and S. C. Shapiro, editors, *Proceedings Sixth International Conference on Principles of Knowledge Representation and Reasoning (KR'98)*, pages 212–223. Morgan Kaufmann Publishers, 1998.

[50] J. McCarthy. Formalization of common sense, papers by John McCarthy edited by V. Lifschitz. Ablex, 1990.

[51] J. McCarthy and P. J. Hayes. Some philosophical problems from the standpoint of artificial intelligence. In B. Meltzer and D. Michie, editors, *Machine Intelligence 4*, pages 463–502. Edinburgh University Press, 1969. reprinted in [50].

[52] D. McDermott. A critique of pure reason. *Computational Intelligence*, 3:151–237, 1987. Cited in [6].

[53] M. W. Moskewicz, C. F. Madigan, Y. Zhao, L. Zhang, and S. Malik. Chaff: Engineering an Efficient SAT Solver. In *Proceedings of the 38th Design Automation Conference, DAC 2001, Las Vegas, NV, USA, June 18-22, 2001*, pages 530–535. ACM, June 2001.

[54] C. H. Papadimitriou. *Computational Complexity*. Addison-Wesley, 1994.

[55] M. A. Peot. *Decision-Theoretic Planning*. PhD thesis, Stanford University, Stanford, CA, USA, 1998.

[56] M. A. Peot and D. E. Smith. Conditional Nonlinear Planning. In *Proceedings of the First International Conference on Artificial Intelligence Planning Systems*, pages 189–197. AAAI Press, 1992.

[57] A. Polleres. The DLV$^{\mathcal{K}}$ System for Planning with Incomplete Knowledge. Master's thesis, Institut für Informationssysteme, Technische Universität Wien, Wien, Österreich, 2001.

[58] L. Pryor and G. Collins. Planning for Contingencies: A Decision-based Approach. *Journal of Artificial Intelligence Research*, 4:287–339, 1996.

[59] R. Reiter. On closed world data bases. In H. Gallaire and J. Minker, editors, *Logic and Data Bases*, pages 55–76. Plenum Press, New York, 1978.

[60] J. Rintanen. Constructing Conditional Plans by a Theorem-Prover. *Journal of Artificial Intelligence Research*, 10:323–352, 1999.

[61] J. Rintanen. Improvements to the evaluation of quantified boolean formulae. In T. Dean, editor, *Proceedings of the Sixteenth International Joint Conference on Artificial Intelligence (IJCAI) 1999*, pages 1192–1197, Stockholm, Sweden, Aug. 1999. Morgan Kaufmann Publishers.

[62] S. J. Russel and P. Norvig. *Artificial Intelligence, A Modern Approach*. Prentice-Hall, Inc., 1995.

[63] M. Shanahan. Prediction is deduction but explanation is abduction. In *Proceedings IJCAI '89*, pages 1055–1060, 1989.

[64] D. E. Smith and D. S. Weld. Conformant Graphplan. In *Proceedings of the Fifteenth National Conference on Artificial Intelligence, (AAAI'98)*, pages 889–896. AAAI Press / The MIT Press, July 1998.

[65] V. Subrahmanian and C. Zaniolo. Relating Stable Models and AI Planning Domains. In L. Sterling, editor, *Proceedings of the 12$^{th}$ International Conference on Logic Programming*, pages 233–247, Tokyo, Japan, June 1995. MIT Press.

[66] G. J. Sussman. The Virtuous Nature of Bugs. In J. Allen, J. Hendler, and A. Tate, editors, *Readings in Planning*, chapter 3, pages 111–117. Morgan Kaufmann Publishers, Inc., 1990. Originally written 1974.

[67] H. Turner. Representing actions in logic programs and default theories: A situation calculus approach. *Journal of Logic Programming*, 31(1–3):245–298, 1997.

[68] H. Turner. A logic of universal causation. *Artificial Intelligence*, 113:87–123, 1999.

[69] H. Turner. Polynomial-length planning spans the polynomial hierarchy. Unpublished manuscript, 2001.

[70] J. D. Ullman. *Principles of Database and Knowledge Base Systems*, volume 1. Computer Science Press, 1989.

[71] M. Veloso. Nonlinear problem solving using intelligent causal-commitment. Technical Report CMU-CS-89-210, Carnegie Mellon University, 1989.

[72] D. S. Weld. An Introduction to Least Commitment Planning. *AI Magazine*, 15(4):27–61, 1994.

[73] D. S. Weld. Recent Advances in AI Planning. *AI Magazine*, 20(2):93–123, 1999.

[74] H. Zhang. SATO: An Efficient Propositional Prover. In *Proceedings of the International Conference on Automated Deduction (CADE'1997)*, pages 272–275, 1997.

# A Appendix: Further Examples of Problem Solving in $\mathcal{K}$

This appendix contains encodings of three well-known planning problems, which should further illustrate the practical use of language $\mathcal{K}$.

## A.1 The Yale Shooting Problem

Another example for dealing with incomplete knowledge is a variation of the famous Yale Shooting Problem (see [32]). We assume here that the agent has a gun and does not know whether it is initially loaded. This can be modeled as follows:

```
fluents :    alive. loaded.
actions :    load. shoot.
always :     executable shoot if loaded.
             executable load if not loaded.
             caused -alive after shoot.
             caused -loaded after shoot.
             caused loaded after load.
initially :  total loaded.
             alive.
goal :       -alive ? (1)
```

The `total` statement leads to two possible legal initial states: $s_1 = \{\texttt{loaded}, \texttt{alive}\}$ and $s_2 = \{-\texttt{loaded}, \texttt{alive}\}$. With $s_1$ `shoot` is executable, while it is not with $s_2$. Executing `shoot` establishes the goal, so the planning problem has the optimistic plan

$$\langle\{\texttt{shoot}\}\rangle$$

which is not secure because of $s_2$.

## A.2 The Monkey and Banana Problem

This example is a variation of the Monkey and Banana problem as described in the CCALC manual (<URL:http://www.cs.utexas.edu/users/mccain/cc/>). It shows that in $\mathcal{K}$ the applicability of actions can be formulated very intuitively by using the `executable` statement. The encoding in CCALC uses many `nonexecutable` statements instead.

In the background knowledge we have three objects: the monkey, the banana and a box.

```
object(box).  object(monkey).  object(banana).
```

Furthermore there are three locations: 1, 2 and 3.

```
location(1).  location(2).  location(3).
```

In the beginning, the monkey is at location 1, the box is at location 2, and the banana is hanging from the ceiling over location 3. The monkey shall get the banana by moving the box towards it, climbing the box, and then grasping the banana hanging from the ceiling. We solve this problem using the following $\mathcal{K}$ program:

```
fluents : at(O,L) requires object(O), location(L).
          onBox.
          hasBanana.
actions : walk(L) requires location(L).
          pushBox(L) requires location(L).
          climbBox.
          graspBanana.
always :  caused at(monkey,L) after walk(L).
          caused -at(monkey,L) after walk(L1), at(monkey,L), L <> L1.
          executable walk(L) if not onBox.
          caused at(monkey,L) after pushBox(L).
          caused at(box,L) after pushBox(L).
          caused -at(monkey,L) after pushBox(L1), at(monkey,L), L <> L1.
          caused -at(box,L) after pushBox(L1), at(box,L), L <> L1.
          executable pushBox(L) if at(monkey,L1), at(box,L1), not onBox.
          caused onBox after climbBox.
          executable climbBox if not onBox, at(monkey,L), at(box,L).
          caused hasBanana after graspBanana.
          executable graspBanana if onBox, at(monkey,L), at(banana,L).
          inertial at(O,L).
          inertial onBox.
          inertial hasBanana.
initially : at(monkey,1).
            at(box,2).
            at(banana,3).
noConcurrency.
goal :      hasBanana ? (4)
```

For this planning problem, the following secure plan exists:

$\langle\{\texttt{walk(2)}\}, \{\texttt{pushBox(3)}\}, \{\texttt{climbBox}\}, \{\texttt{graspBanana}\}\rangle$

Let us now deal with incomplete knowledge about the location of objects. Similar as in the Blocks World example in Section 3.2, we introduce a new fluent:

```
objectIsSomewhere(O) requires object(O).
```

Furthermore, we add the following constraints and rules in the initial state:

```
forbidden at(O,L), at(O,L1), L <> L1.
forbidden onBox, at(monkey,L), notatBox(L).
caused objectIsSomewhere(O) if at(O,L).
forbidden not objectIsSomewhere(O).
```

These constraints guarantee a correct initial state.

## A.3 The Rocket Transport Problem

This example is a variation of a planning problem for rockets introduced in [71]. There are two one-way rockets, which can transport cargo objects from one place to another. The objects have to be loaded on the rocket and unloaded at the destination. This example shows the capability of $\mathcal{K}$ to deal with concurrent actions, as the two rockets can be loaded, can move, and can be unloaded in parallel.

The background knowledge consists of three places, the two rockets and the objects to transport:

```
rocket(sojus). rocket(apollo).
cargo(food). cargo(tools). cargo(car).
place(earth). place(mir). place(moon).
```

The action description for the rocket planning domain comprises three actions `move(R,L)`, `load(C,R)` and `unload(C,R)`. The fluents are `atR(R,L)` (where the rocket currently is), `atC(C,L)` (where the cargo object currently is), `in(C,R)` (describing that an object is inside a rocket) and `hasFuel(R)` (the rocket has fuel and can move). Now let us solve the problem of transporting the car to the moon and food and tools to Mir, given that all objects are initially on the earth and both rockets have fuel. We define the following planning problem:

```
fluents:   atR(R,P) requires rocket(R), place(P).
           atC(C,P) requires cargo(C), place(P).
           in(C,R) requires rocket(R), cargo(C).
           hasFuel(R) requires rocket(R).
actions:   move(R,P) requires rocket(R), place(P).
           load(C,R) requires rocket(R), cargo(C).
           unload(C,R) requires rocket(R), cargo(C).
always:    caused atR(R,P) after move(R,P).
           caused -atR(R,P) after move(R,P1), atR(R,P).
           caused -hasFuel(R) after move(R,P).
           executable move(R,P) if hasFuel(R), not atR(R,P).
           caused in(C,R) after load(C,R).
           caused -atC(C,P) after load(C,R), atC(C,P).
           executable load(C,R) if atC(C,P), atR(R,P).
           caused atC(C,P) after unload(C,R), atR(R,P).
           caused -in(C,R) after unload(C,R).
           executable unload(C,R) if in(C,R).
           nonexecutable move(R,P) if load(C,R).
           nonexecutable move(R,P) if unload(C,R).
           nonexecutable move(R,P) if move(R,P1), P <> P1.
```

```
             nonexecutable load(C,R) if load(C,R1), R <> R1.
             inertial atC(C,L).
             inertial atR(R,L).
             inertial in(C,R).
             inertial hasFuel(R).
initially :  atR(R,earth).
             atC(C,earth).
             hasFuel(R).
goal :       atC(car,moon), atC(food,mir), atC(tools,mir) ? (3)
```

The `nonexecutable` statements exclude simultaneous actions as follows:

- loading/unloading a rocket and moving it;
- moving a rocket to two different places;
- loading an object on two different rockets.

For the given goal, there are two secure plans, where in the first one rocket `sojus` flies to the moon and `apollo` flies to Mir, and in the second one the roles are interchanged:

```
⟨ {load(food,sojus), load(tools,sojus), load(car,apollo)},
  {move(sojus,mir), move(apollo,moon)},
  {unload(food,sojus), unload(tools,sojus), unload(car,apollo)} ⟩
⟨ {load(car,sojus), load(food,apollo), load(tools,apollo)},
  {move(sojus,moon), move(apollo,mir)},
  {unload(car,sojus), unload(food,apollo), unload(tools,apollo)} ⟩
```